\documentclass{article} 
\usepackage{iclr2027_conference,times}

\usepackage{amsmath,amsfonts,bm}

\def\eqref#1{equation~\ref{#1}}

\def\1{\bm{1}}

\DeclareMathAlphabet{\mathsfit}{\encodingdefault}{\sfdefault}{m}{sl}
\SetMathAlphabet{\mathsfit}{bold}{\encodingdefault}{\sfdefault}{bx}{n}

\usepackage{hyperref}
\usepackage{url}
\usepackage{graphicx}
\usepackage{amsthm}
\usepackage{booktabs}
\usepackage{multirow}
\usepackage{wrapfig}
\usepackage{subcaption}
\usepackage{pifont}

\newcommand{\cmark}{\ding{51}}
\newcommand{\xmark}{\ding{55}}

\newtheorem{proposition}{Proposition}

\title{From Suppression to Repair: Mitigating Object Hallucination in Large Vision-Language Models via Localized Distribution Alignment}

\author{
\begin{minipage}[t]{\dimexpr\textwidth-2\tabcolsep\relax}
\centering
\normalfont\normalsize
Chen Zhao$^{1}$ \quad
Xingping Dong$^{1,\dagger}$ \quad
Jiachun Shi$^{1}$ \quad
Liang Peng$^{1}$ \\[3pt]
Chong Wang$^{1}$ \quad
Zhen Lei$^{2}$ \quad
Ran He$^{2}$ \quad
Bo Du$^{1}$ \\[8pt]
{\small
$^{1}$School of Computer Science, Wuhan University\\
$^{2}$Institute of Automation, Chinese Academy of Sciences
}\\[6pt]
{\footnotesize
\texttt{zhaochen.whu2026@gmail.com} \quad
\texttt{xingpingdong@whu.edu.cn}\\
\texttt{2023302111051@whu.edu.cn} \quad
\texttt{pengliang@whu.edu.cn}\\
\texttt{cwang@whu.edu.cn} \quad
\texttt{zhen.lei@ia.ac.cn}\\
\texttt{rhe@nlpr.ia.ac.cn} \quad
\texttt{dubo@whu.edu.cn}
}\\[6pt]
\end{minipage}
}

\iclrfinalcopy 
\begin{document}

\maketitle
\pagestyle{plain}
\thispagestyle{plain}

\raggedbottom
\setlength{\textfloatsep}{12pt plus 2pt minus 2pt}

\begin{abstract}
Object hallucination remains a major obstacle for large vision-language models (LVLMs) to generate reliable content. An intuitive mitigation strategy is to suppress hallucination-related components in hidden representations. However, these components may also contain useful information, and suppressing them can weaken the model's multimodal capabilities. In this paper, we propose ResOT, a training-free method that repairs representations at inference time through localized distribution alignment. Specifically, ResOT projects dominant hallucinated directions away from the faithful subspace, forming a low-dimensional residual subspace for intervention. Within this subspace, ResOT uses Gaussian optimal transport (OT) to align the hallucinated distribution with the faithful one. The resulting map defines repair targets with minimal changes to the original representations. At inference, ResOT adaptively controls how far each token state moves toward its OT target. Experiments on three representative LVLMs show that ResOT substantially reduces object hallucination while improving image caption quality and multimodal performance across multiple benchmarks. Code will be released.
\end{abstract}

\section{Introduction}
Large vision-language models (LVLMs) can generate fluent and informative responses to visual input~\citep{bai2023qwenvl, chen2024internvl}, but they still describe objects that are not present in the image. This phenomenon, known as object hallucination, undermines the reliability of LVLMs~\citep{bai2024hallusurvey, li2023pope, rohrbach2018chair}. One approach to reducing object hallucination is to fine-tune LVLMs with additional data, but this often incurs substantial training costs~\citep{jiang2024hacl, liu2024lrvinstruction}. Representation-based methods offer a training-free alternative by modifying the model's hidden representations~\citep{liu2025vti, lin2026hulluedit}. Some of these methods identify hallucination-related directions or subspaces and suppress the corresponding components~\citep{yang2025nullu, dastmalchi2026cipher}. However, \textbf{being associated with hallucination does not make a component specific to hallucination}~\citep{hua2025sae, zhu2026mpd}. The same components may also carry useful information for visual understanding, so suppressing them can compromise image caption quality or multimodal capabilities.

\begin{figure}[t]
    \centering
    \includegraphics[width=1.0\linewidth]{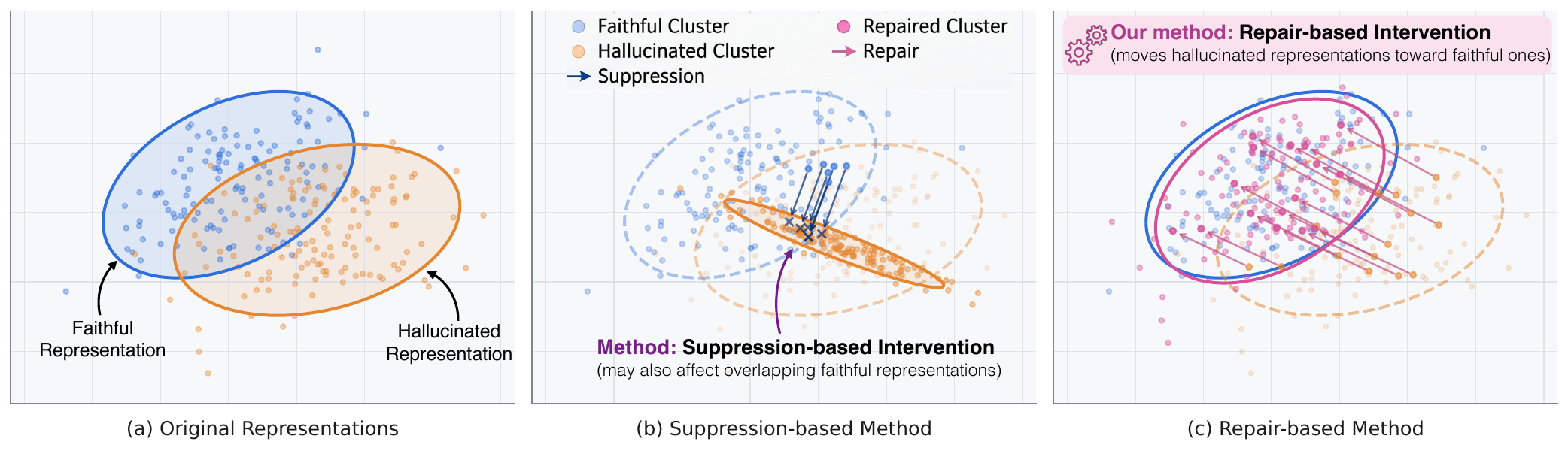}
    \caption{
    \textbf{Conceptual comparison of suppression and repair.} (a) Faithful (blue) and hallucinated (orange) representations form partially overlapping clusters. (b) Suppression compresses the orange cluster into a narrow band and also displaces some faithful points (blue arrows). (c) Repair instead moves the orange points toward the blue cluster (pink arrows), forming the repaired cluster in pink.
    }
    \label{fig:suppression_vs_repair}
\end{figure}

Therefore, we seek to repair the hallucination-related components rather than discard them. Instead of simply reducing the contribution of these components, we use their distribution in faithful representations to set repair targets. Figure~\ref{fig:suppression_vs_repair} illustrates the key distinction: suppression-based interventions reduce the variation of hallucinated representations along the selected direction, but may also inadvertently alter faithful representations; in contrast, repair-based interventions use the faithful distribution to define where the representation should move. 

The choice of alignment space is also important, as it determines which distributional differences the repair addresses. The dominant faithful subspace is spanned by the leading directions of faithful representations. We examine a low-dimensional residual subspace orthogonal to it. Our held-out analysis shows that this residual subspace reveals a clearer mismatch between faithful and hallucinated representations (Section~\ref{section_4_4}). Importantly, faithful representations are not confined to their dominant subspace; they still have components in the residual subspace. Removing these components would discard variation present in faithful representations. These observations motivate localized distribution alignment: \textbf{expose and correct the residual mismatch while leaving components in the dominant faithful subspace unchanged.}

In this paper, we introduce \textbf{ResOT, a training-free method that implements this localized repair at inference time.} During offline calibration, ResOT processes each image with its paired faithful and hallucinated descriptions to collect two sets of hidden representations. It estimates the dominant subspaces of these representations, then projects the dominant hallucinated directions away from the faithful subspace. The leading residual directions define the repair subspace. We project both sets of calibration representations into this subspace and fit a Gaussian distribution to each. Gaussian optimal transport (OT) then provides a closed-form map from the hallucinated distribution to the faithful one. During inference, ResOT projects the current token state into the same subspace and applies the map to obtain its repair target. A likelihood-based gate compares the projected state's likelihood under the two fitted distributions and controls how far it moves toward the target. The resulting correction is added back to the hidden state along the residual directions, leaving the orthogonal component unchanged. The LVLM weights remain fixed throughout calibration and inference.

Our theoretical analysis supports the repair subspace and transport target. The residual subspace captures as much of the dominant hallucinated directions as possible while remaining orthogonal to the dominant faithful subspace, helping expose the residual differences to be repaired. In this subspace, if  components are removed via orthogonal projection, suppression cannot match the original faithful distribution as long as those components are also present in the faithful representation. Gaussian OT instead aligns the fitted residual distributions with minimum expected squared displacement.

We combined ResOT with LLaVA-1.5, mPLUG-Owl2, and InstructBLIP~\citep{liu2023llava, ye2024mplug, dai2023instructblip} and conducted experiments on multiple benchmarks. The results demonstrate that ResOT reduces object hallucination while improving image caption quality and multimodal performance. Compared to the vanilla models, it reduces the CHAIR$_{s}$ metric by 13.9\%–46.8\% and achieves average relative improvements of 27.67\% and 5.86\% on the BLEU-1 and MME metrics~\citep{papineni2002bleu, fu2023mme}, respectively. On LLaVA-1.5, ResOT achieves the highest throughput among the evaluated inference-time methods that keep the LVLM weights unchanged. It retains 81.24\% of the original model's generation throughput (tokens per second) while requiring only an additional 8.15 MiB of peak GPU memory. 

\section{Related Work}
\subsection{Hallucination in Large Vision-Language Models}
Object hallucination in LVLMs has been studied from multiple perspectives, such as visual perception, cross-modal alignment, and reliance on language priors~\citep{gu2025marine, jain2024vcoder, xing2024cca, liu2024payatten}. One line of research involves training-based methods, which typically use supervised fine-tuning~\citep{liu2024lrvinstruction}, contrastive learning~\citep{jiang2024hacl}, RLHF~\citep{sun2024mmhalbench}, and DPO~\citep{yang2025dpo, yu2024rlhfv}. However, these methods often require additional supervision signals and parameter optimization. Other approaches focus on interventions during decoding. They adjust token selections through visual contrast~\citep{leng2024vcd, chen2024halc}, attention signals~\citep{zhou2025causalmm, huang2024opera}, or intermediate-layer predictions~\citep{wang2025deco, chuang2024dola}. Although these methods avoid retraining, they can introduce substantial inference overhead due to extra forward passes or complex decoding procedures. Post-hoc approaches revise unsupported content using learned revisors~\citep{zhou2024lure} or visual verification~\citep{yin2024woodpecker}. These methods add a correction stage after generation.

ResOT is most closely related to representation-based interventions. Existing methods steer latent activations~\citep{liu2025vti, li2025vista}, intervene on attention heads~\citep{chen2025ict}, or suppress hallucination-related components~\citep{yang2025nullu, dastmalchi2026cipher}. More selective approaches use evidence-consistent subspace editing~\citep{lin2026hulluedit} or targeted parameter updates~\citep{zhu2026mpd} to limit interference with useful information. ResOT instead defines a faithful target distribution within a residual subspace. Its key distinction is the use of distribution alignment rather than component removal to determine the update.

\subsection{Optimal Transport for Representation Alignment}
Optimal transport (OT) aligns distributions by finding a transport plan that minimizes the cost of moving probability mass between them~\citep{villani2009optimal, peyre2019computational}. It has been widely used for feature alignment and domain adaptation~\citep{courty2017optimal, damodaran2018deepjdot}. Empirical OT estimates a coupling between samples, often using iterative solvers such as Sinkhorn~\citep{cuturi2013sinkhorn}. Under squared Euclidean cost, the OT map between non-degenerate Gaussian distributions has a closed-form affine expression determined by their means and covariances~\citep{peyre2019computational}. ResOT uses this map to define repair targets in a low-dimensional residual subspace. The map is estimated offline and applied during generation, without an online transport solver.

\section{Method}
\label{main:method}
\begin{figure}[t]
    \centering
    \includegraphics[width=1.0\linewidth]{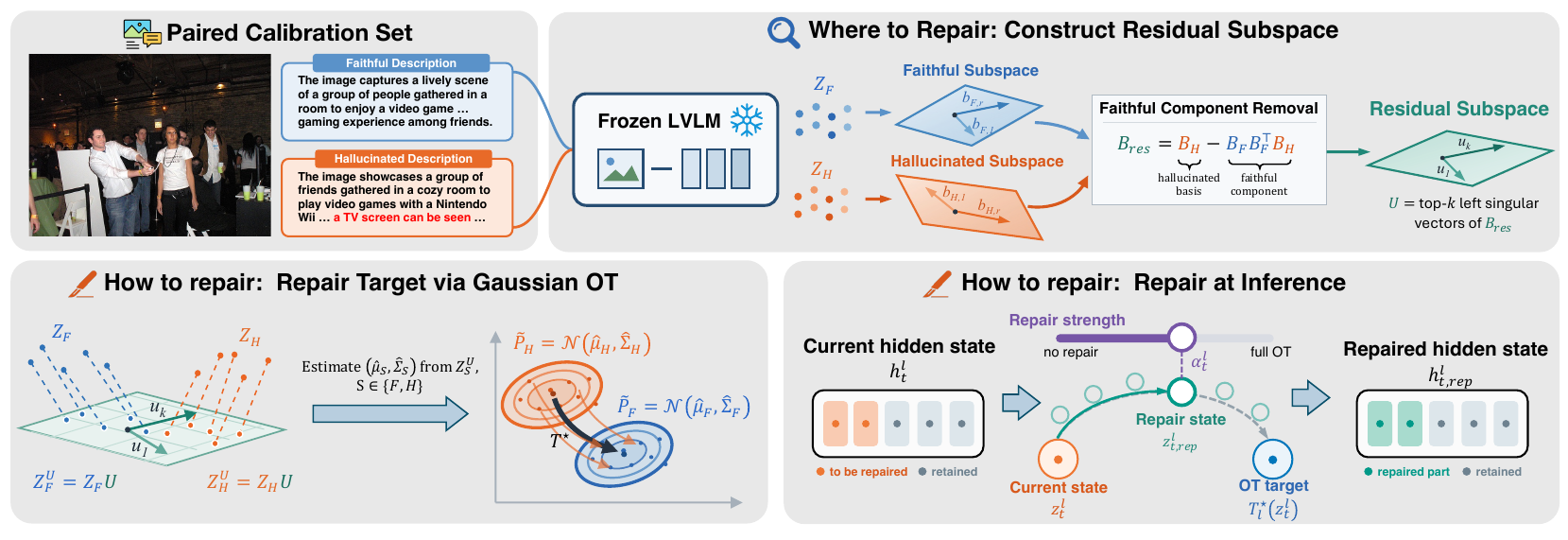}
    \caption{
        \textbf{Overview of ResOT}. \textbf{Where to repair (top)}: the orange hallucinated directions are projected away from the blue faithful subspace to form the green residual subspace. \textbf{How to repair (bottom):} Gaussian OT aligns the orange and blue residual distributions to define repair targets (left); a likelihood-based gate controls the movement toward each target during inference (right). 
    }
    \label{fig:resot_overview}
\end{figure}

ResOT consists of two stages: offline calibration and online repair, as shown in Figure~\ref{fig:resot_overview}. During calibration, a residual subspace is identified at each selected layer, and a Gaussian optimal transport map is estimated within this subspace. During inference, ResOT uses this map to define a repair target for the current token state and adaptively controls the movement toward that target.

\subsection{LVLM Generation and Calibration Representations}
An LVLM encodes an image $\mathcal{V}$ into visual features and maps them to the input space of the language decoder~\citep{liu2023llava, dai2023instructblip, ye2024mplug}. By combining these visual features with the prompt text $x$, the decoder can generate a response autoregressively. At decoding step $t$, the available context consists of the image, the prompt, and the previously generated tokens $y_{<t}$. We denote by $h_t^\ell\in\mathbb{R}^d$ the hidden state at the final text position in layer $\ell$, computed from this context (i.e., the state prior to the generation of $y_t$).

Like prior representation-based interventions~\citep{yang2025nullu, zhu2026mpd}, ResOT uses a paired calibration set to determine where to repair and how the repair should be performed. The set contains $N$ images, each paired with a faithful description and a hallucinated description. We process the two image–description inputs separately and extract the hidden state at the final text position of each complete description. At each selected layer, stacking these states gives $Z_F^\ell, Z_H^\ell\in\mathbb{R}^{N\times d}$. We refer to them as faithful and hallucinated representations according to the descriptions from which they are extracted.

ResOT estimates its subspaces and transport maps from these complete descriptions and reuses them at the current decoding position. This follows the strategy of deriving textual steering directions from final-position representations of paired captions and applying them during generation~\citep{liu2025vti}. During the calibration and inference stages, the parameters of the LVLM remain fixed.

\subsection{Residual Subspace Representation Repair}

\subsubsection{Locating the Repair Subspace}
ResOT constructs a repair subspace from the dominant hallucinated directions that lie outside the dominant faithful subspace. This step projects the hallucinated basis to select intervention directions. For clarity, we omit the layer index $\ell$ when describing offline calibration.

Given $Z_F, Z_H\in\mathbb{R}^{N\times d}$, we compute their singular value decompositions: $Z_F = L_F S_F R_F^{\top}$, $Z_H = L_H S_H R_H^{\top}$. With the right singular vectors ordered by the decreasing singular value, we use the top-$r$ columns of $R_F$ and $R_H$ to form $B_F=[b_{F,1},b_{F,2},\ldots,b_{F,r}]\in\mathbb{R}^{d\times r}$ and $B_H=[b_{H,1},b_{H,2},\ldots,b_{H,r}]\in\mathbb{R}^{d\times r}$, where $B_F$ and $B_H$ are orthonormal bases for the dominant faithful and hallucinated subspaces. The projection of $B_H$ onto the faithful subspace is $B_{F}B_F^{\top}B_H$. Removing this component gives
\begin{equation}
    B_{\mathrm{res}} = B_{H}-B_FB_F^{\top}B_{H}
                     = (I_{d}-B_FB_F^{\top})B_H.
\end{equation}
We compute $B_{\mathrm{res}}=\bar{L}\bar{S}\bar{R}^\top$ and let $q=\operatorname{rank}(B_{\mathrm{res}})$. For $k\leq q$, the residual basis is $U=\bar{L}[:,1:k]\in\mathbb{R}^{d\times k}$. Its columns are orthonormal and satisfy $B_F^\top U=0$. All subsequent updates lie in $\operatorname{span}(U)$, leaving components in the dominant faithful subspace unchanged. However, faithful representations are not confined to their dominant subspace and can still have nonzero components in the residual subspace. We therefore use the distribution of these components as a reference for repair rather than setting all residual components to zero.

\subsubsection{Gaussian Optimal Transport in the Residual Subspace}
After obtaining the residual subspace, we project both sets of calibration representations into the residual subspace: $Z_F^U=Z_FU$, $Z_H^U=Z_HU$, where each row is a $k$-dimensional residual coordinate vector. Let $(\hat{\mu}_F,\hat{\Sigma}_F)$ and $(\hat{\mu}_H,\hat{\Sigma}_H)$ denote their empirical means and covariances. Critically, a finite calibration set may produce noisy or ill-conditioned covariance estimates. We therefore apply covariance shrinkage~\citep{ledoit2004covshrink} before computing the transport map:
\begin{equation}
    \hat{\Sigma}_s \leftarrow (1-\rho)\hat{\Sigma}_s + \rho\frac{\mathrm{tr}(\hat{\Sigma}_s)}{k}I_k+\epsilon I_k, \qquad s\in\{F,H\}.
\end{equation}
Here, $\rho\in[0,1]$ is the shrinkage coefficient, and $\epsilon>0$ ensures positive definiteness. We then approximate the residual distributions as $\tilde{P}_F = \mathcal{N}(\hat{\mu}_F, \hat{\Sigma}_F)$ and $\tilde{P}_H = \mathcal{N}(\hat{\mu}_H, \hat{\Sigma}_H)$.

We seek a map that transports the hallucinated approximation to the faithful one with minimum expected squared displacement:
\begin{equation}
    T^{\star}=\arg\min_{T\in\mathcal{T}}\mathbb{E}_{z\sim \tilde{P}_H}\big\|T(z)-z\big\|_2^2 \quad \mathrm{subject \ to} 
    \quad 
    T_{\#}\tilde{P}_H =\tilde{P}_F.
\end{equation}
Here, $\mathcal{T}$ denotes measurable maps from $\mathbb{R}^k$ to $\mathbb{R}^k$, and $T_{\#}\tilde{P}_H$ is the distribution obtained by applying $T$ to samples from $\tilde{P}_H$. For non-degenerate Gaussian distributions, the optimal map has a closed-form affine solution~\citep{brenier1991polar, peyre2019computational}:
\begin{equation}
T^\star(z) = \hat{\mu}_F + A(z-\hat{\mu}_H),
\end{equation}
where $A=\hat{\Sigma}_H^{-\frac{1}{2}}\left(\hat{\Sigma}_H^{\frac{1}{2}}\hat{\Sigma}_F\hat{\Sigma}_H^{\frac{1}{2}}\right)^{\frac{1}{2}}\hat{\Sigma}_H^{-\frac{1}{2}}$. The map exactly aligns the fitted Gaussian distributions and defines a full OT target $T^{\star}(z)$ for each residual state $z$. Since $U$ has orthonormal columns,
\begin{equation}
    \left\| U(T^{\star}(z)-z)\right\|_2=\left\| T^{\star}(z)-z\right\|_2.
\end{equation}
Thus, the displacement measured in residual coordinates equals the magnitude of the corresponding correction in the original hidden space.

\subsubsection{Repair at Inference}
At inference, ResOT decomposes the current hidden state into residual and orthogonal components. For the residual component, the OT map defines a full repair target, while a likelihood-based gate determines how much of the correction is applied. Subsequently, this repaired residual component is combined with the unchanged orthogonal component to form the repaired hidden state.

Each selected layer $\ell$ has a residual basis $U^{\ell}$ and an OT map $T_{\ell}^{\star}$, both estimated offline. At decoding step $t$, we compute
\begin{equation}
    z_t^\ell=(U^\ell)^\top h_t^\ell,\qquad h_{t,\perp}^\ell=\bigl(I_d-U^\ell(U^\ell)^\top\bigr)h_t^\ell.
\end{equation}
The full OT target is $\tilde{z}_t^\ell=T_\ell^\star(z_t^\ell)$. The OT map defines where the residual state should move, but it does not determine how strongly the full correction should be applied to each decoding state. We therefore use a likelihood-based gate to adjust the fraction of the correction applied to each state:
\begin{equation}
     s_t^{\ell}=\frac{1}{k}\left[\log\mathcal{N}\left(z_t^{\ell};\hat\mu_H^{\ell},\hat\Sigma_H^{\ell}\right)-\log\mathcal{N}\left(z_t^{\ell};\hat\mu_F^{\ell},\hat\Sigma_F^{\ell}\right)\right],\qquad g_t^{\ell}=\sigma(s_t^{\ell}),
\end{equation}
where $\sigma$ is the sigmoid function. A large $g_t^{\ell}$ indicates stronger relative evidence for the hallucinated approximation and assigns a larger fraction of the OT correction. We set $\alpha_t^{\ell}=\alpha g_t^{\ell}$, where $\alpha\in[0,1]$ is the global repair strength. We then interpolate between the current residual state and its OT target:
\begin{equation}
    z_{t,\mathrm{rep}}^{\ell}=(1-\alpha_t^{\ell})z_t^{\ell} + \alpha_t^{\ell}\tilde{z}_t^{\ell}.
\end{equation}
Combining this coordinate with the unchanged orthogonal component, we have
\begin{equation}
    h_{t,\mathrm{rep}}^{\ell} = h_t^{\ell} + \alpha_t^{\ell} U^{\ell}(T^{\star}_{\ell}({z}_t^{\ell}) - z_t^{\ell}).
\end{equation}
In practice, we additionally clip the norm of the OT correction to prevent unusually large updates. The repaired state then proceeds through the remaining frozen decoder layers.

\subsubsection{Theoretical Analysis}
We next analyze ResOT's residual subspace construction and Gaussian OT, and compare them with projection-based suppression methods, which we model as a rank-reducing orthogonal projection. All proofs are provided in Appendix~\ref{proof}.

\begin{proposition}
Let $B_{\mathrm{res}}=(I_d-B_FB_F^{\top})B_H$, and $U\in\mathbb{R}^{d \times k}$ contain its top-$k$ left singular vectors, where $k \leq \operatorname{rank}(B_{\mathrm{res}})$. For any orthonormal basis $V\in\mathbb{R}^{d\times k}$ satisfying $B_F^{\top}V=0$,
$$
\left\| V^{\top}B_H \right\|_F^2\leq \left\| U^{\top}B_H \right\|_F^2=\sum_{i=1}^k\sigma_i^2(B_{\mathrm{res}}).
$$
\end{proposition}
This term $\left\|V^{\top}B_H \right\|_F^2$ measures the total squared projection of the dominant hallucinated directions onto a candidate subspace. ResOT maximizes this for a fixed dimension, while excluding the dominant faithful subspace.

\begin{proposition}
Let $\tilde{P}_H, \tilde{P}_F\in\mathcal{P}_2(\mathbb{R}^k)$, where $\mathcal{P}_2(\mathbb{R}^k)$ denotes probability distributions with finite second moments. Let $\mu_F$ and $\Sigma_F$ be the mean and covariance of $\tilde{P}_F$. For $T_{\Pi}(z)=\Pi z$, where $\Pi=\Pi^{\top}=\Pi^{2}$ and $\mathrm{rank}(\Pi)<k$,
$$
W_2^2\big((T_\Pi)_\# \tilde{P}_H, \tilde{P}_F\big)=W_2^2\big((T_{\Pi})_{\#}\tilde{P}_H, (T_{\Pi})_{\#}\tilde{P}_F\big)+\big\|(I_k - \Pi)\mu_F\big\|_2^2 + \operatorname{tr}\big((I_k - \Pi)\Sigma_F\big).
$$
\end{proposition}
The first term measures the degree of mismatch within the retained subspace. The remaining terms equal the expected squared norm of the faithful component that was removed by projection. If this value is positive, even when the projected distribution matches perfectly, there will still be a positive Wasserstein distance from the original faithful distribution.

\begin{proposition}
Let $Z\in\mathbb{R}^{k}$ be any random vector with finite second moment, mean $\mu_H$ and covariance $\Sigma_{H}\in\mathbb{S}^{k}_{++}$. Given a target $\mu_F\in\mathbb{R}^k$ and $\Sigma_F\in\mathbb{S}_{++}^k$, define
$$
\mathcal{A}_{H\rightarrow F}=\left\{T_A(z)=\mu_F+A(z-\mu_H)\mid A\Sigma_H A^{\top}=\Sigma_F \right\}.
$$
Then the Gaussian OT map $T^{\star}$ is the unique minimizer:
$$
T^\star=\arg\min_{T_A\in\mathcal{A}_{H\rightarrow F}}\mathbb{E}\left[||T_A(Z)-Z||_2^2\right].
$$
\end{proposition}
This objective depends only on the first two moments of the source distribution. Therefore, the same affine map remains optimal without assuming that the source distribution is Gaussian. When both the source and target distributions are Gaussian, this map also aligns their complete distributions; otherwise, the proposition guarantees only the optimality of the affine moment matching.

\section{Experiments}
We organize our evaluation around three questions. 
\textbf{RQ1}: Does ResOT reduce object hallucination while maintaining caption quality?
\textbf{RQ2}: Does ResOT preserve multimodal performance?
\textbf{RQ3}: How do ResOT's residual subspace and OT-based repair affect task performance, and how robust is the method to calibration size and corruption?

\subsection{Experimental Setup}
\paragraph{Baselines.}
ResOT is calibrated using 5,000 paired descriptions from the LURE training data~\citep{zhou2024lure}. We compare ResOT against vanilla LVLMs and two categories of training-free mitigation methods.
The vanilla baselines are LLaVA-1.5~\citep{liu2023llava}, mPLUG-Owl2~\citep{ye2024mplug}, and InstructBLIP~\citep{dai2023instructblip}, all equipped with 7B language decoders and evaluated using greedy decoding. Decoding-time baselines include OPERA~\citep{huang2024opera}, VCD~\citep{leng2024vcd}, HALC~\citep{chen2024halc}, and SGRS~\citep{zhang2026lvlm_latency}. Representation-level baselines include ICT~\citep{chen2025ict}, VTI~\citep{liu2025vti}, VISTA~\citep{li2025vista}, Nullu~\citep{yang2025nullu}, HulluEdit~\citep{lin2026hulluedit}, and Revis~\citep{wu2026revis}.

\paragraph{Benchmarks and Metrics.}
We evaluate ResOT along three dimensions: object hallucination, generation quality, and multimodal performance. POPE~\citep{li2023pope} measures discriminative object hallucination through yes/no object-existence questions. We report accuracy, precision, and F1 score. CHAIR~\citep{rohrbach2018chair} evaluates hallucinated objects in open-ended captions using CHAIR$_s$ ($\text{C}_s$) and CHAIR$_i$ ($\text{C}_i$), while BLEU-1~\citep{papineni2002bleu} evaluates caption quality. MME~\citep{fu2023mme} evaluates multimodal perception and cognition. We further use GPT-4o to assess the accuracy and detail of open-ended descriptions on sampled COCO images~\citep{lin2014coco}. To broaden the evaluation, we assess hallucination on MMHal-Bench and AMBER~\citep{sun2024mmhalbench, wang2023amber}. We also evaluate multimodal capabilities on ScienceQA, MMBench, TextVQA, MMMU, and MM-Vet~\citep{lu2022learn, liu2023mmbench, singh2019textvqa, yue2024mmmu, yu2023mmvet}. Full results are reported in Appendix~\ref{app:additional_experimental_results}.

\subsection{Object Hallucination and Caption Quality (RQ1)}
To answer RQ1, we evaluate ResOT in two settings. POPE measures discriminative object hallucination, while CHAIR measures hallucination in open-ended image captioning. We also assess caption quality using BLEU-1 and a GPT-based evaluation of visual accuracy and detailedness.

\begin{table}[t]
\centering
\small
\setlength{\tabcolsep}{4.2pt}
\caption{Average POPE results across the random, popular and adversarial splits. Higher values indicate better performance. The best results are highlighted in bold.}
\label{tab:pope_results}
\renewcommand{\arraystretch}{1.08}
\begin{tabular}{lccccccccc}
\toprule
\multirow{2}{*}{\textbf{Method}}
& \multicolumn{3}{c}{\textbf{LLaVA-1.5}}
& \multicolumn{3}{c}{\textbf{mPLUG-Owl2}}
& \multicolumn{3}{c}{\textbf{InstructBLIP}} \\
\cmidrule(lr){2-4} \cmidrule(lr){5-7} \cmidrule(lr){8-10}
& Acc. $\uparrow$ & Prec. $\uparrow$ & F1 $\uparrow$
& Acc. $\uparrow$ & Prec. $\uparrow$ & F1 $\uparrow$
& Acc. $\uparrow$ & Prec. $\uparrow$ & F1 $\uparrow$ \\
\midrule
Vanilla   & 84.20 & 82.30 & 84.67 & 78.23 & 72.07 & 81.07 & 84.17 & 83.07 & 84.60 \\
OPERA     & 84.67 & 83.37 & 85.13 & 79.77 & 74.37 & 82.03 & 85.00 & 83.93 & 85.43 \\
VCD       & 81.13 & 77.83 & 82.30 & 76.73 & 70.83 & 79.77 & 75.97 & 78.20 & 81.63 \\
HALC      & 84.83 & 82.80 & 85.53 & 78.30 & 72.23 & 81.13 & 84.60 & 83.37 & 85.13 \\
VISTA     & 84.23 & 80.03 & 85.47 & 76.00 & 69.33 & 79.70 & 85.20 & 84.87 & 85.50 \\
ICT       & 85.43 & 83.10 & 86.07 & 79.07 & 73.37 & 81.57 & 55.50 & 53.00 & 69.20 \\
VTI       & 85.21 & 81.73 & 86.23 & 80.09 & 79.75 & 81.96 & 83.70 & 81.00 & 84.63 \\
Nullu     & 85.53 & \textbf{86.07} & 85.70 & 77.87 & 71.60 & 80.87 & 84.63 & 83.20 & 85.20 \\
Revis     & 85.19 & 82.15 & 86.09 & 78.24 & 72.10 & 81.12 & 84.82 & 83.78 & 85.30 \\ 
HulluEdit & 85.63 & 82.76 & 86.46 & 79.64 & 74.59 & 81.72 & 83.08 & 80.25 & 84.14 \\
SGRS & 85.12 & 81.96 & 86.05 & 78.01 & 71.78 & 80.98 & 84.46 & 83.09 & 85.03 \\

\midrule
ResOT    & \textbf{86.13} & 84.43 & \textbf{86.54}
         & \textbf{82.50} & \textbf{81.57} & \textbf{82.49}
         & \textbf{86.15} & \textbf{86.93} & \textbf{85.94} \\
\bottomrule
\end{tabular}
\end{table}
\paragraph{Results on POPE.}
Table~\ref{tab:pope_results} reports results averaged over the random, popular, and adversarial splits. 
ResOT achieves the highest accuracy and F1 score across all three backbones. It also obtains the highest precision on mPLUG-Owl2 and InstructBLIP. These improvements show that ResOT improves object existence judgments across the three LVLMs.

\begin{table}[t]
\caption{Captioning results on COCO evaluated by CHAIR and BLEU-1. Lower $\text{C}_s$ and $\text{C}_i$ indicate fewer hallucinated objects, while higher BLEU-1 indicates better caption quality. The maximum generation length is set to 512 tokens. The best results are highlighted in bold.}
\label{tab:chair_bleu_results}
\centering
\footnotesize
\setlength{\tabcolsep}{4pt}
\renewcommand{\arraystretch}{1.08}
\resizebox{0.90\linewidth}{!}{
\begin{tabular}{lccccccccc}
\toprule
\multirow{2}{*}{\textbf{Method}}
& \multicolumn{3}{c}{\textbf{LLaVA-1.5}}
& \multicolumn{3}{c}{\textbf{mPLUG-Owl2}}
& \multicolumn{3}{c}{\textbf{InstructBLIP}} \\
\cmidrule(lr){2-4} \cmidrule(lr){5-7} \cmidrule(lr){8-10}
& $\text{C}_s$ $\downarrow$ & $\text{C}_i$ $\downarrow$ & BLEU-1 $\uparrow$
& $\text{C}_s$ $\downarrow$ & $\text{C}_i$ $\downarrow$ & BLEU-1 $\uparrow$
& $\text{C}_s$ $\downarrow$ & $\text{C}_i$ $\downarrow$ & BLEU-1 $\uparrow$ \\
\midrule
Vanilla
& 0.514 & 0.1451 & 0.1739
& 0.568 & 0.1639 & 0.1559
& 0.462 & 0.1310 & 0.1740 \\

OPERA
& 0.454 & 0.1317 & 0.1769
& 0.488 & 0.1503 & 0.1769
& \textbf{0.398} & 0.1242 & 0.1524 \\


VCD
& 0.484 & 0.1437 & 0.1743
& 0.580 & 0.1808 & 0.1689
& 0.566 & 0.1808 & 0.1698 \\

HALC
& 0.380 & 0.1270 & 0.1870
& 0.500 & 0.1432 & 0.1762
& 0.620 & 0.1770 & 0.1580 \\

VISTA
& 0.436 & 0.1336 & 0.1619
& 0.528 & 0.1638 & 0.1426
& 0.434 & 0.1345 & 0.1508 \\

ICT
& 0.498 & 0.1439 & 0.1797
& 0.716 & 0.2037 & 0.1585
& 0.482 & 0.1349 & 0.1196 \\

VTI
& 0.322 & 0.1195 & 0.2017
& 0.314 & 0.1194 & 0.2087
& 0.428 & 0.1236 & 0.1379 \\

Nullu
& 0.394 & 0.1150 & 0.1581
& 0.412 & 0.1395 & 0.1830
& 0.440 & 0.1429 & 0.1658 \\

Revis
& 0.430 & 0.1361 & 0.1740
& 0.460 & 0.1489 & 0.1711
& 0.440 & 0.1211 & 0.1744\\

HulluEdit
& 0.404 & 0.1287 & 0.1669
& 0.362 & 0.1324 & 0.1515
& 0.442 & 0.1203 & 0.1633\\

SGRS
& 0.348 & 0.1132 & 0.2091
& 0.330 & 0.1188 & 0.1798
& 0.436 & 0.1288 & 0.1757 \\

\midrule
ResOT
& \textbf{0.292} & \textbf{0.0953} & \textbf{0.2334}
& \textbf{0.302} & \textbf{0.0898} & \textbf{0.2293}
& \textbf{0.398} & \textbf{0.1146} & \textbf{0.1770} \\
\bottomrule
\end{tabular}
}
\end{table}
\paragraph{Results on CHAIR.}
Table~\ref{tab:chair_bleu_results} evaluates open-ended COCO captions. ResOT achieves the lowest $\mathrm{C}_s$ and $\mathrm{C}_i$. These reductions are accompanied by higher BLEU-1 scores, rather than a decline in reference-based image description quality. Table~\ref{tab:caption_quality} reports additional captioning metrics.

\begin{table}[t]
\caption{GPT-4o evaluation on COCO image descriptions. Each entry reports the baseline/ResOT scores for visual accuracy (Acc.) and detailedness (Det.); higher values indicate better performance.}
\label{tab:gpt_coco_eval}
\centering
\footnotesize
\setlength{\tabcolsep}{4.5pt}
\renewcommand{\arraystretch}{1.05}
\begin{tabular}{@{}lcccc@{}}
\toprule
\multirow{2}{*}{\textbf{Backbone}} 
& \multicolumn{2}{c}{\textbf{Vanilla vs. ResOT}} 
& \multicolumn{2}{c}{\textbf{Nullu vs. ResOT}} \\
\cmidrule(lr){2-3} \cmidrule(lr){4-5}
& \textbf{Acc.} & \textbf{Det.} & \textbf{Acc.} & \textbf{Det.} \\
\midrule
LLaVA-1.5 
& 5.72 / \textbf{6.77} 
& \textbf{6.09} / 5.81 
& 6.22 / \textbf{6.32} 
& 5.76 / \textbf{5.82} \\
mPLUG-Owl2 
& 5.65 / \textbf{6.52} 
& 5.90 / \textbf{5.94} 
& 6.38 / \textbf{6.88} 
& 6.22 / \textbf{6.24} \\
InstructBLIP 
& 5.68 / \textbf{6.38} 
& 5.82 / \textbf{5.94} 
& 5.36 / \textbf{6.48} 
& 5.72 / \textbf{6.00} \\
\bottomrule
\end{tabular}
\end{table}
\paragraph{GPT-based Evaluation.} 
We use GPT-4o to compare descriptions on the 500 COCO validation images selected by OPERA. Following the evaluation prompt used in previous works~\citep{wang2025deco,yang2025nullu}, we repeat the scoring five times and report the mean scores. We compare ResOT with the vanilla backbone and Nullu, a representative suppression-based method. As shown in Table~\ref{tab:gpt_coco_eval}, ResOT achieves higher accuracy than both baselines across all three backbones and higher detailedness in five of the six comparisons. These results demonstrate improved visual faithfulness while retaining informative content.

\subsection{Multimodal Capability Assessment (RQ2)}
To answer RQ2, we use MME to evaluate multimodal performance, which covers both perception and cognition. Additional experimental results are reported in Appendix~\ref{app:amber_mmhal_mmvet_results}.

\begin{figure}[t]
\centering
\includegraphics[width=\textwidth]{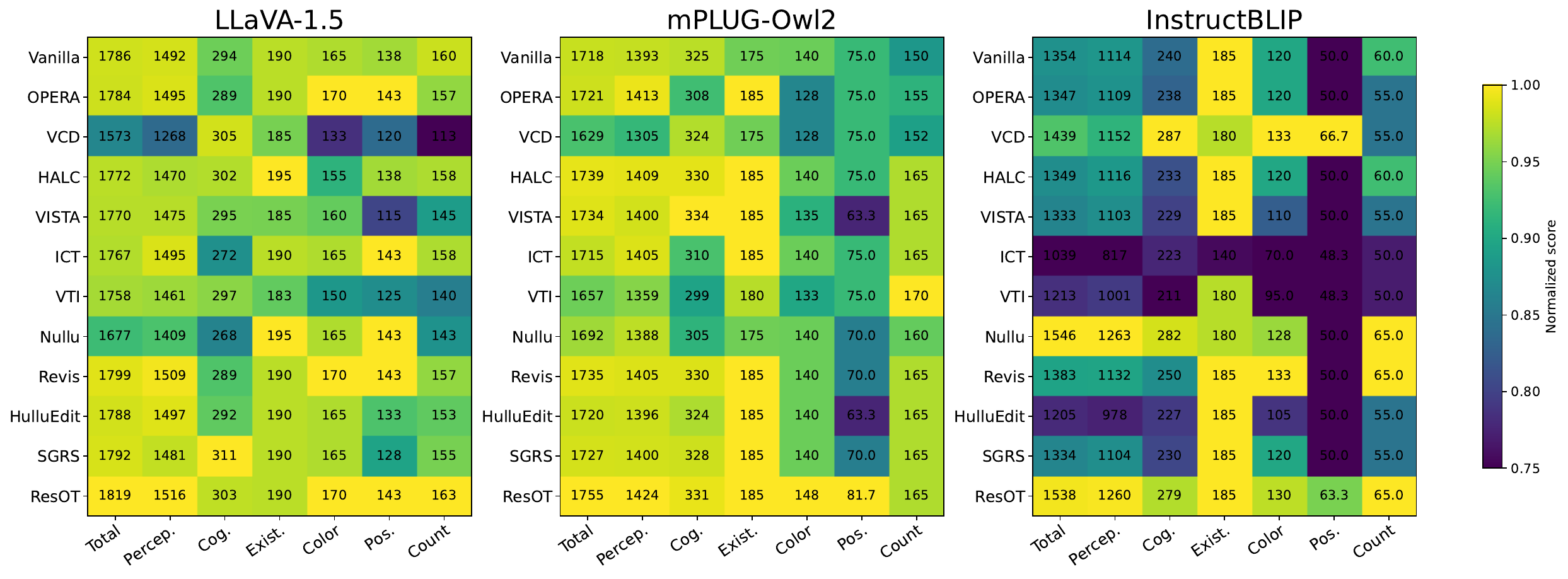}
\caption{MME results across three LVLM backbones. Each cell reports the raw score, while color intensity indicates the normalized score within each metric. We report the total, perception and cognition scores, together with four perception subcategories: existence, color, position and count.}
\label{fig:mme_heatmap}
\end{figure}

\paragraph{Results on multimodal capability benchmarks.}
Figure~\ref{fig:mme_heatmap} reports MME results. ResOT improves the total, perception, and cognition scores over the vanilla model on all three backbones. It achieves the highest total and perception scores on LLaVA-1.5 and mPLUG-Owl2, together with the best or tied-best results in most object-related subcategories. 
We further experiment on other multimodal capability benchmarks. Across MMMU, TextVQA, ScienceQA, and MMBench, ResOT outperformed Vanilla in 9 out of 12 settings (Table~\ref{tab:multimodal_benchmarks}). These results demonstrate that ResOT preserves and often improves broader multimodal performance.

\subsection{Analysis of Localized Representation Repair}
\label{section_4_4}
To answer RQ3, we compare ResOT with Nullu via representation visualizations and ablations of the intervention subspace and repair operation. We also evaluate calibration sample efficiency and robustness to contamination. More mechanism analyses are provided in Appendix~\ref{app:sens_and_abla_analysis}.

\begin{figure}[t]
\centering
\includegraphics[width=1.0\textwidth]{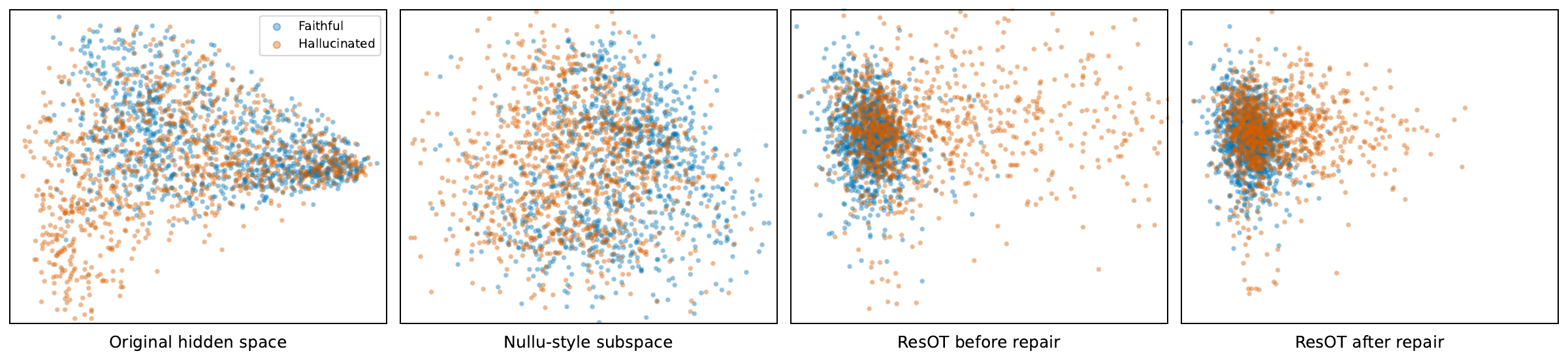}
\caption{\textbf{Representation visualization at layer 24.} From left to right: the original representation space, the Nullu-style subspace, and the ResOT residual subspace before and after repair. Faithful and hallucinated representations are evaluated on the held-out split. We use PCA for visualization.}
\label{fig:projection_scatter_layer31}
\end{figure}

\paragraph{Representation Distribution Visualization.}
In our held-out analysis, we use 4,000 calibration pairs to estimate the parameters, while the remaining 1,000 pairs are used for visualization and quantitative analysis. Figure~\ref{fig:projection_scatter_layer31} visualizes the representations at layer 24. We can see that the residual subspace exposes a clearer difference between faithful and hallucinated representations than the original and Nullu-style subspaces. Since there is still overlap between these two distributions, the residual subspace should not be interpreted as containing only hallucination-specific features. OT repair reduces the mismatch without discarding the residual components. This provides empirical evidence for the distinction between localization and suppression.

\paragraph{Ablation Study.}
\begin{wraptable}{r}{0.54\textwidth}
\vspace{-9pt}
\caption{Ablation study results on subspace construction and intervention operations.}
\label{tab:ablation}
\centering
\footnotesize
\setlength{\tabcolsep}{3.2pt}
\renewcommand{\arraystretch}{1.05}

\begin{tabular}{lcccc}
\toprule
\textbf{Variant}
& $\text{C}_s$ $\downarrow$
& BLEU-1 $\uparrow$
& MME $\uparrow$ \\
\midrule
RandSub + OT
& 0.516 & 0.1748 & 1787.17 \\
NulluSub + OT
& 0.424 & 0.1681 & 1764.82 \\
ResSub + Sup
& 0.378 & 0.1903 & 1784.96 \\
ResSub + Mean
& 0.320 & 0.2140 & 1798.35 \\
ResSub + OT w/o Gate
& \textbf{0.292}  & 0.2268 & 1814.17 \\
ResSub + OT w/ Gate
& \textbf{0.292} & \textbf{0.2334} & \textbf{1819.02} \\
\bottomrule
\end{tabular}
\end{wraptable}
Table~\ref{tab:ablation} separates the effects of the intervention subspace, repair operation, and likelihood-based gate. We first compare subspaces using ungated Gaussian OT. The residual subspace achieves lower $\mathrm{C}_s$ and higher BLEU-1 and MME than either the random or Nullu-style subspace, showing that the choice of repair subspace is crucial. 
With the residual subspace fixed, replacing suppression with ungated Gaussian OT achieves better results. Mean-only alignment improves over suppression but remains below Gaussian OT on all three metrics. Therefore, matching covariance as well as the mean provides an additional benefit.
At the same reported $\mathrm{C}_s$, adding the likelihood-based gate further improves BLEU-1 and MME.

\begin{figure*}[t]
    \centering
    \includegraphics[width=\textwidth]{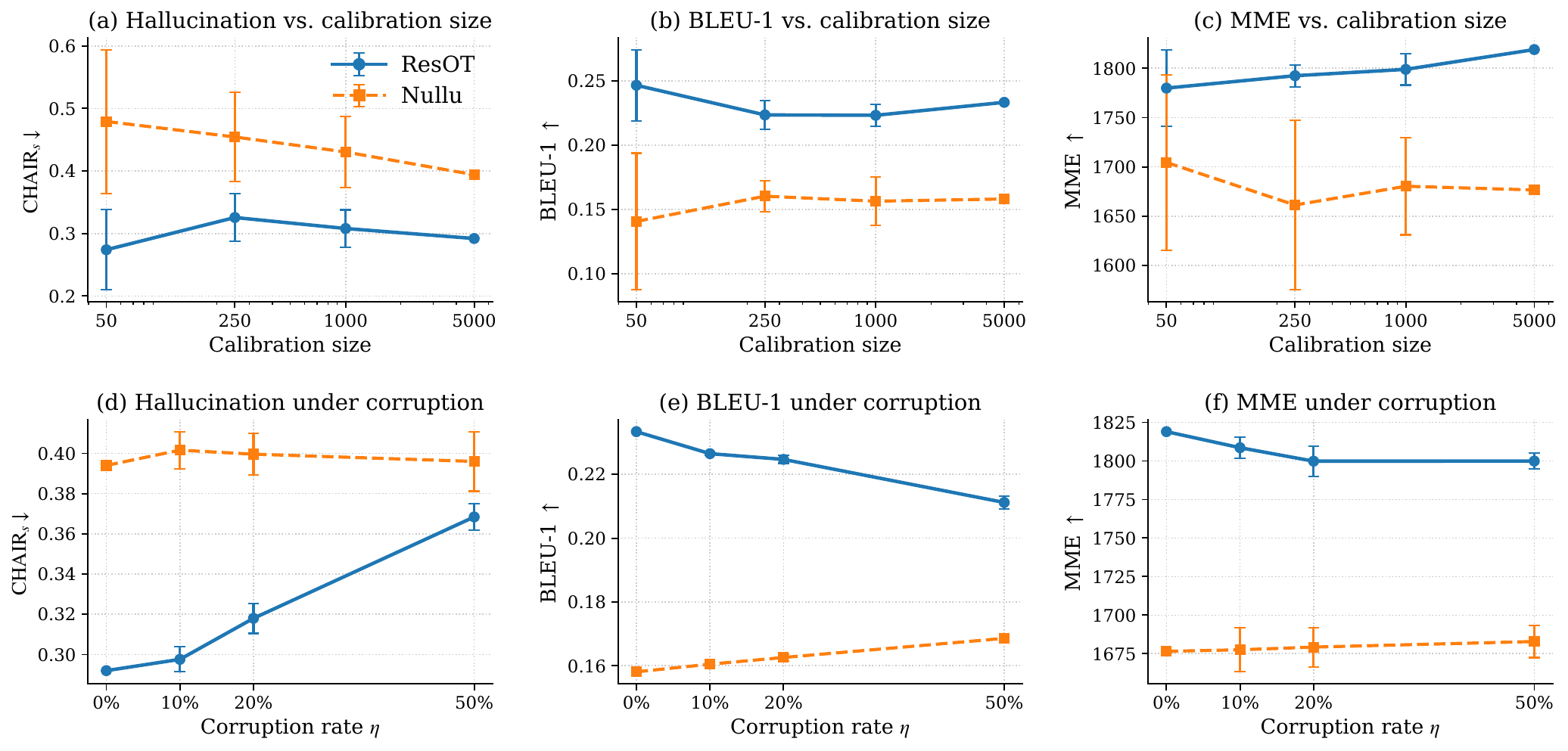}
    \caption{
    Effect of calibration size and corruption. We compare ResOT and Nullu under different calibration sizes and corruption rates. Error bars denote the standard deviation across random trials.
    }
    \label{fig:calibration_analysis}
\end{figure*}

\paragraph{Calibration Set Analysis.}
Figure~\ref{fig:calibration_analysis} examines the effect of calibration size and corruption. In experiments controlling for the number of calibration pairs, ResOT already outperforms Nullu in $\mathrm{C}_s$, BLEU-1, and MME with only 50 calibration pairs. Performance variability across trials generally decreases as the calibration set grows. We also corrupt an $\eta$ fraction of the calibration pairs, with the corrupted samples split evenly between image--text mismatches and faithful/hallucinated label swaps. As the corruption rate $\eta$ increases, ResOT retains its advantage over Nullu on all three metrics. These results demonstrate that ResOT remains effective with few calibration pairs and is robust to imperfect calibration data.

\section{Conclusion}
In this work, we introduce ResOT, which mitigates object hallucination by repairing hallucination-related representations rather than directly suppressing them. ResOT confines the intervention to the residual subspace and uses Gaussian optimal transport to move representations toward a faithful target with minimum expected squared displacement. A likelihood-based gate further adjusts how much of this correction is applied during decoding. Across multiple benchmarks and three representative LVLMs, ResOT consistently reduces object hallucination while preserving or even improving image caption quality and broader multimodal performance. These results indicate that hallucination-related components do not always need to be removed; repairing their distribution provides a practical alternative to direct suppression. 



\bibliography{iclr2027_conference}
\bibliographystyle{iclr2027_conference}

\appendix
\section{Proof}
\label{proof}
\subsection{Proof of Proposition 1}
Since $B_F$ has orthonormal columns,
$$
B_F^{\top}B_{\mathrm{res}}=B_{F}^{\top}(I_d-B_FB_F^{\top})B_H=0.
$$
Therefore, every left singular vector of $B_{\mathrm{res}}$ associated with a nonzero singular value is orthogonal to $\operatorname{Span}(B_F)$. Since $k \leq \operatorname{rank}(B_{\mathrm{res}})$, the selected basis $U$ satisfies $B_F^{\top}U=0$.

Now consider any feasible orthonormal basis $V$ such that $B_F^{\top}V=0$. This condition gives $V^{\top}B_FB_F^{\top}=0$, and hence, 
$$
V^{\top}B_H=V^{\top}(I_d-B_FB_F^{\top})B_H=V^{\top}B_{\mathrm{res}}.
$$
Let $B_{\mathrm{res}}=Q\Sigma R^{\top}$ be a full singular value decomposition with $Q=\left[q_1,...,q_d\right]$ and $\sigma_1 \geq \cdots \geq \sigma_d \geq 0$. Then
$$
\left\|V^{\top}B_H\right\|_F^2=
\left\|V^{\top}B_\mathrm{res}\right\|_F^2=\sum_{i=1}^{d}\sigma_{i}^2\left\|V^{\top}q_i\right\|_2^2.
$$
Define $a_i=\left\| V^{\top}q_i\right\|_2^2$. Since $V^{\top}V=I_k$ and $Q$ is orthogonal, we have
$$
\sum_{i=1}^d a_i=\operatorname{tr}(V^{\top}QQ^{\top}V)=k, \qquad 0 \leq a_i \leq 1.
$$
The singular values are ordered in decreasing order, so the weighted sum is maximized by placing unit weight on the first $k$ terms:
$$
\sum_{i=1}^d\sigma_i^2a_i\le \sum_{i=1}^k\sigma_i^2.
$$
This upper bound is attained by $V=U=[q_1,...,q_k]$. Since $U^{\top}B_H=U^{\top}B_{\mathrm{res}}$, we obtain
$$
\left\| V^{\top}B_H \right\|_F^2\leq \left\| U^{\top}B_H \right\|_F^2=\sum_{i=1}^k\sigma_i^2(B_{\mathrm{res}}).
$$
This proves the claim. \hfill$\square$

\subsection{Proof of Proposition 2}
Let $Q_H=(T_{\Pi})_{\#}\tilde{P}_H$ and $Q_F=(T_{\Pi})_{\#}\tilde{P}_F$, where $T_{\Pi}(z)=\Pi z$. Consider any coupling $\gamma \in \Pi(Q_{H}, \tilde{P}_{F})$ and let $(X,Y)\sim \gamma$. 
Because $X$ lies in the range of $\Pi$, we have $\Pi X=X$ and $(I_k-\Pi)X=0$. The retained and removed components are orthogonal, so
$$
\begin{aligned}
||X-Y||_2^2
& = ||\Pi(X-Y)||_2^2 + ||(I_k-\Pi)(X-Y)||_2^2 \\
& = ||\Pi(X-Y)||_2^2+||(I_k-\Pi)Y||_2^2. \\
\end{aligned}
$$
The joint distribution of $(X, \Pi Y)$ is a coupling of $Q_H$ and $Q_F$. Therefore,
$$
\mathbb{E}_{\gamma}\left[||X-\Pi Y||_2^2\right] \geq W_2^2(Q_H,Q_F).
$$
The second term depends only on the marginal distribution of $Y$, which gives
$$
\mathbb{E}_{\gamma}\left[||(I_k-\Pi)Y||_2^2\right]=\mathbb{E}_{Y\sim\tilde{P}_F}[||(I_k-\Pi)Y||_2^2].
$$
It follows that
$$
W_2^2(Q_H, \tilde{P}_F)\geq W_2^2(Q_H, Q_F) + \mathbb{E}_{Y\sim\tilde{P}_F}[||(I_k-\Pi)Y||_2^2].
$$
It remains to show that this lower bound is attainable. Let $\eta$ be an optimal coupling between $Q_H$ and $Q_F$. Sample $(X,U)\sim \eta$, and then sample $Y$ from the conditional distribution of $\tilde{P}_F$ given $\Pi Y=U$. This gives a coupling between $Q_H$ and $\tilde{P}_F$ for which $(X,\Pi Y)\sim \eta$. Hence, 
$$
\begin{aligned}
\mathbb{E}\left[||X-Y||_2^2\right] 
& = \mathbb{E}\left[||X-\Pi Y||_2^2\right] + \mathbb{E}\left[||(I_k-\Pi)Y||_2^2\right] \\
& = W_2^2(Q_H,Q_F) + \mathbb{E}_{Y\sim\tilde{P}_F}\left[||(I_k-\Pi)Y||_2^2\right].
\end{aligned}
$$
Combining the two inequalities gives
$$
W_2^2(Q_H,\tilde{P}_F)=W_2^2(Q_H,Q_F)+\mathbb{E}_{Y\sim\tilde{P}_F}\left[||(I_k-\Pi)Y||_2^2\right].
$$
Let $\mu_F$ and $\Sigma_F$ be the mean and covariance of $\tilde{P}_F$. The remaining expectation can be written as
$$
\begin{aligned}
\mathbb{E}\left[||(I_k-\Pi)Y||_2^2\right]
& = ||(I_k-\Pi)\mu_F||_2^2 + \operatorname{tr}\big((I_k-\Pi)\Sigma_F(I_k-\Pi)\big) \\
& = ||(I_k-\Pi)\mu_F||_2^2 + \operatorname{tr}\big((I_k-\Pi)\Sigma_F\big).
\end{aligned}
$$
Substituting this expression gives
$$
\begin{aligned}
W_2^2\big((T_\Pi)_\# \tilde{P}_H, \tilde{P}_F\big) 
& = W_2^2\big((T_{\Pi})_{\#}\tilde{P}_H, (T_{\Pi})_{\#}\tilde{P}_F\big) \\
& +\big\|(I_k - \Pi)\mu_F\big\|_2^2 + \operatorname{tr}\big((I_k - \Pi)\Sigma_F\big).
\end{aligned}
$$
Thus, whenever the removed subspace contains nonzero faithful energy, the projected distribution
remains at a strictly positive Wasserstein distance from $\tilde{P}_F$. \hfill$\square$

\subsection{Proof of Proposition 3}
Let $X=Z-\mu_H$ and $\delta=\mu_F-\mu_H$. Then $\mathbb{E}\left[X\right]=0$, $\operatorname{Cov}(X)=\Sigma_H$, and
$$
T_{A}(Z)-Z=\delta+(A-I_k)X.
$$
Since the cross term has zero expectation,
$$
\begin{aligned}
\mathbb{E}\left[||T_A(Z)-Z||_2^2\right]
& = ||\delta||^2_2 + \mathbb{E}\left[||(A-I_k)X||_2^2\right] \\
& = ||\delta||_2^2 + \operatorname{tr}\left[(A-I_k)\Sigma_H(A-I_k)^{\top}\right] \\
& = || \mu_F-\mu_H||_2^2 + \operatorname{tr}(\Sigma_H) + \operatorname{tr}(A\Sigma_H A^{\top})-2\operatorname{tr}(A\Sigma_H)\\
& = ||\mu_F-\mu_H||_2^2 + \operatorname{tr}(\Sigma_H)+\operatorname{tr}(\Sigma_F)-2\operatorname{tr}(A\Sigma_H). \\
\end{aligned}
$$
The first three terms are independent of $A$. Thus, minimizing the expected distortion is equivalent to solving
$$
\max\limits_{A:A\Sigma_H A^{\top}=\Sigma_F}\operatorname{tr}(A\Sigma_H).
$$
For any feasible $A$, define $Q=\Sigma_F^{-\frac{1}{2}}A\Sigma_H^{\frac{1}{2}}$.
The feasibility constraint implies
$$
QQ^{\top}=\Sigma_F^{-\frac{1}{2}}A\Sigma_HA^{\top}\Sigma_F^{-\frac{1}{2}}=I_k.
$$
So $Q$ is orthogonal. We can write $A=\Sigma_F^{\frac{1}{2}}Q\Sigma_H^{-\frac{1}{2}}$, which gives $\operatorname{tr}(A\Sigma_H)=\operatorname{tr}(Q\Sigma_H^{\frac{1}{2}}\Sigma_F^{\frac{1}{2}})$. Let $M=\Sigma_H^{\frac{1}{2}}\Sigma_F^{\frac{1}{2}}$.
By the von Neumann trace inequality, for any orthogonal $Q$,
$$
\operatorname{tr}(QM)\le ||M||_{\star}=\operatorname{tr}\left((MM^{\top})^{\frac{1}{2}}\right).
$$
Since $MM^{\top} = \Sigma_H^{\frac{1}{2}}\Sigma_F\Sigma_H^{\frac{1}{2}}$, we have
$$
\operatorname{tr}(A\Sigma_H)\leq \operatorname{tr}\left[(\Sigma_H^{\frac{1}{2}}\Sigma_F\Sigma_H^{\frac{1}{2}})^{\frac{1}{2}}\right].
$$
We define $S=(\Sigma_{H}^{\frac{1}{2}}\Sigma_{F}\Sigma_{H}^{\frac{1}{2}})^{\frac{1}{2}}$. The proposed matrix $A^{\star}=\Sigma_H^{-\frac{1}{2}}S\Sigma_H^{-\frac{1}{2}}$ is feasible. Furthermore,
$$
\operatorname{tr}(A^{\star}\Sigma_H)=\operatorname{tr}(\Sigma_H^{-\frac{1}{2}}S\Sigma_H^{\frac{1}{2}}) = \operatorname{tr}(S).
$$
Therefore, $A^{\star}$ attains the upper bound and therefore minimizes the expected displacement.
It remains to show that the minimizer is unique. Let $M=U\Lambda V^{\top}$ be a singular value decomposition of $M$. Since $\Sigma_H$ and $\Sigma_F$ are positive definite, every singular value in $\Lambda$ is strictly positive. For any orthogonal $Q$,
$$
\operatorname{tr}(QM)=\operatorname{tr}(V^{\top}QU\Lambda)\leq \operatorname{tr}(\Lambda).
$$
The equality holds only when $V^{\top}QU=I_k$. Hence, the corresponding feasible $A$ is unique. Since $A^{\star}$ attains the maximum, it is the unique minimizer. Consequently, $T^{\star}$ is the unique affine map that matches the first and second moments of the target with the smallest expected squared change, even when the source distribution is non-Gaussian. \hfill$\square$

\section{Implementation and Evaluation Details}
\paragraph{Backbones and Calibration Data.}
We evaluate ResOT on LLaVA-1.5-7B, InstructBLIP-Vicuna-7B, and mPLUG-Owl2-LLaMA2-7B. For the vanilla models and ResOT, we use greedy decoding. For each backbone, we use the same 5,000 calibration pairs from the LURE training data~\citep{zhou2024lure}. Each pair contains an image from the COCO training set, a faithful caption from LLaVA-Instruct-150K, and a corresponding hallucinated caption generated by GPT-3.5.

\paragraph{Evaluation Protocols.}
For POPE, we evaluate on the random, popular, and adversarial splits, each containing 3,000 samples. We parse each response by checking whether its first generated word is ``yes'' or ``no''; otherwise, the prediction is labeled as unknown and counted as incorrect.
For CHAIR, we evaluate object hallucination on the COCO 2014 validation set~\citep{lin2014coco}. Following prior work~\citep{huang2024opera, yang2025nullu}, we use the subset of 500 images selected by OPERA for evaluation. We use the fixed prompt: \textit{Please describe this image in detail}.
For MME, we follow the official yes/no evaluation protocol. Each image is paired with two questions, and the score combines question-level accuracy with image-level pair accuracy.
For GPT-assisted open-ended evaluation, we use GPT-4o as an external judge to assess image description quality. Given an image, the prompt and two model responses, GPT-4o assigns each response a score from 1 to 10 for visual accuracy and detailedness. We repeat each comparison five times and report the mean score.

\paragraph{Baseline Implementation.}
We compare with decoding-based methods, including OPERA, VCD, HALC, and SGRS, and representation-level methods, including ICT, VTI, VISTA, Nullu, HulluEdit, and Revis. For each baseline, we use the official code and the hyperparameters recommended in the corresponding paper. Vanilla and ResOT use greedy decoding. OPERA uses beam search with five beams, while SGRS uses saliency-guided top-k rejection sampling. Within each benchmark, we keep the LVLM backbone, prompt, and maximum generation length the same across all methods.

\paragraph{Hyperparameter Settings.}
ResOT uses two rank parameters. The rank $r$ determines the number of dominant directions used to estimate the faithful and hallucinated subspaces, while $k$ determines the dimension of the residual repair subspace. We set $r=k$ and select this shared rank from $\{8,16,32,64,128\}$. The repair strength $\alpha$ is selected from $\{0.1,0.2,\ldots,1.0\}$. At each decoding step, repair is applied only to the final-position hidden state in decoder layers 16--31. We set the covariance shrinkage coefficient to $\rho=0.05$ and the numerical stability term to $\epsilon=10^{-6}$. For each layer, the correction threshold is set to the 95th percentile of OT correction norms computed from the hallucinated calibration representations.

We split the calibration set by image into fitting and validation subsets of 4,500 and 500 pairs, respectively. The fitting subset is used to estimate the residual subspaces and Gaussian OT maps, while the validation subset is used to select the shared rank $r=k$ and repair strength $\alpha$. For each candidate configuration, we measure generative hallucination using $\mathrm{CHAIR}_i$ based on COCO ground-truth object annotations, object-level multimodal preservation using balanced accuracy on yes/no questions constructed from present and absent objects, and generation quality using BLEU-1 against the faithful reference captions. We first retain the five configurations with the highest balanced accuracy, and then select the one with the best average rank across $\mathrm{CHAIR}_i$ and BLEU-1. After selecting $r=k$ and $\alpha$, we estimate the residual subspaces and Gaussian OT maps using all 5,000 calibration pairs.

\section{Additional Experimental Results}
\label{app:additional_experimental_results}
\subsection{More Evaluation of Caption Quality}
\label{app:more_caption_metric}
\begin{table*}[t]
\centering
\caption{Comparison of caption quality across different LVLMs. Higher values are better.}
\label{tab:caption_quality}
\resizebox{\textwidth}{!}{
\begin{tabular}{@{}llccccccc@{}}
\toprule
Model & Method
& BLEU-1
& BLEU-2
& BLEU-3
& BLEU-4
& ROUGE-L
& CLIPScore
& SPICE \\
\midrule

\multirow{3}{*}{LLaVA-1.5}
& Vanilla
& 0.1739 & 0.1174 & 0.0738 & 0.0465
& 0.1804 & 0.8174 & 0.1827 \\
& Nullu
& 0.1581 & 0.1091 & 0.0689 & 0.0429
& 0.1707 & 0.8215 & 0.1761 \\
& ResOT
& \textbf{0.2334} & \textbf{0.1554}
& \textbf{0.0979} & \textbf{0.0609}
& \textbf{0.2062} & \textbf{0.8251}
& \textbf{0.1961} \\
\midrule

\multirow{3}{*}{mPLUG-Owl2}
& Vanilla
& 0.1559 & 0.1157 & 0.0740 & 0.0472
& 0.1788 & 0.8180 & 0.1783 \\
& Nullu
& 0.1830 & 0.1243 & 0.0788 & 0.0496
& 0.1915 & 0.8217 & 0.1926 \\
& ResOT
& \textbf{0.2293} & \textbf{0.1565}
& \textbf{0.1014} & \textbf{0.0654}
& \textbf{0.2213} & \textbf{0.8281}
& \textbf{0.2105} \\
\midrule

\multirow{3}{*}{InstructBLIP}
& Vanilla
& 0.1740 & 0.1224 & 0.0790 & 0.0505
& 0.1779 & 0.8151 & 0.1884 \\
& Nullu
& 0.1658 & 0.1162 & 0.0747 & 0.0475
& 0.1762 & 0.8139 & 0.1859 \\
& ResOT
& \textbf{0.1770} & \textbf{0.1241}
& \textbf{0.0803} & \textbf{0.0514}
& \textbf{0.1849} & \textbf{0.8180}
& \textbf{0.1887} \\
\bottomrule
\end{tabular}
}
\end{table*}

Across the methods reported in Table~\ref{tab:caption_quality}, ResOT achieves the highest scores on BLEU-1–4~\citep{papineni2002bleu}, ROUGE-L~\citep{lin2004rouge}, CLIPScore~\citep{hessel2021clipscore} and SPICE~\citep{anderson2016spice} for all three backbones. The gains are largest on LLaVA-1.5 and mPLUG-Owl2, while InstructBLIP shows smaller but consistent improvements. These results indicate that ResOT preserves and generally improves generation quality across different LVLMs.

\subsection{Results on Additional Benchmarks}
We further evaluate ResOT on additional hallucination and multimodal benchmarks. AMBER~\citep{wang2023amber} provides fine-grained analysis of object hallucination, while MMHal-Bench~\citep{sun2024mmhalbench} evaluates hallucination in open-ended responses. We also use MM-Vet~\citep{yu2023mmvet}, MMMU~\citep{yue2024mmmu}, TextVQA~\citep{singh2019textvqa}, ScienceQA~\citep{lu2022learn}, and MMBench~\citep{liu2023mmbench} to assess broader multimodal capabilities.

\label{app:amber_mmhal_mmvet_results}
\begin{table}[t]
\centering
\footnotesize
\setlength{\tabcolsep}{5pt}
\renewcommand{\arraystretch}{1.05}
\caption{AMBER results across three LVLM backbones. Lower CHAIR, Hal, and Cog indicate less hallucination, while higher Cover indicates better object coverage.}
\label{tab:amber_results}
\begin{tabular}{llcccc}
\toprule
\textbf{Backbone} & \textbf{Method} 
& \textbf{CHAIR} $\downarrow$ 
& \textbf{Cover} $\uparrow$ 
& \textbf{Hal} $\downarrow$ 
& \textbf{Cog} $\downarrow$ \\
\midrule
\multirow{3}{*}{LLaVA-1.5}
& Vanilla & 7.6 & 48.7 & 35.4 & 4.2 \\
& Nullu    & 7.6 & \textbf{49.2} & 37.8 & 3.7 \\
& ResOT    & \textbf{6.0} & 48.2 & \textbf{25.1} & \textbf{2.4} \\
\midrule
\multirow{3}{*}{mPLUG-Owl2}
& Vanilla & 9.3 & \textbf{50.3} & 40.8 & 5.3 \\
& Nullu    & 6.6 & 48.3 & 27.7 & 3.0 \\
& ResOT    & \textbf{5.0} & 48.9 & \textbf{21.3} & \textbf{1.8} \\
\midrule
\multirow{3}{*}{InstructBLIP}
& Vanilla & 7.5 & 52.8 & 36.3 & 3.8 \\
& Nullu    & 8.6 & 51.3 & 33.9 & 3.6 \\
& ResOT    & \textbf{6.4} & \textbf{52.9} & \textbf{31.1} & \textbf{3.2} \\
\bottomrule
\end{tabular}
\end{table}
\paragraph{Results on AMBER.}
On AMBER, ResOT achieved the lowest CHAIR, Hal, and Cog scores on all three backbones (Table~\ref{tab:amber_results}). Its object coverage remains close to the best result for each backbone. Therefore, the reduction in hallucinations is accompanied by a slight change in the object coverage.

\begin{table}[t]
\centering
\footnotesize
\setlength{\tabcolsep}{4.5pt}
\renewcommand{\arraystretch}{1.03}
\caption{MMHal-Bench results across three LVLM backbones. Avg. denotes average score, and Hall. denotes hallucination rate.}
\label{tab:mmhal_results}
\begin{tabular}{lcccccc}
\toprule
\multirow{2}{*}{\textbf{Method}} 
& \multicolumn{2}{c}{\textbf{LLaVA-1.5}}
& \multicolumn{2}{c}{\textbf{mPLUG-Owl2}}
& \multicolumn{2}{c}{\textbf{InstructBLIP}} \\
\cmidrule(lr){2-3} \cmidrule(lr){4-5} \cmidrule(lr){6-7}
& \textbf{Avg.} $\uparrow$ & \textbf{Hall.} $\downarrow$
& \textbf{Avg.} $\uparrow$ & \textbf{Hall.} $\downarrow$
& \textbf{Avg.} $\uparrow$ & \textbf{Hall.} $\downarrow$ \\
\midrule
Vanilla
& 2.1250 & 0.6458
& 1.9479 & 0.6667
& 1.8229 & 0.6458 \\

Nullu
& 2.1146 & 0.6562
& 1.9583 & \textbf{0.6458}
& 1.6250 & 0.6979 \\

ResOT
& \textbf{2.2708} & \textbf{0.5938}
& \textbf{2.0104} & \textbf{0.6458}
& \textbf{1.9583} & \textbf{0.6146} \\
\bottomrule
\end{tabular}
\end{table}
\paragraph{Results on MMHal-Bench.}
On MMHal-Bench, ResOT achieves the highest average score on all three backbones and reduces the hallucination rate compared to Vanilla (Table~\ref{tab:mmhal_results}). It obtains the lowest hallucination rate on LLaVA-1.5 and InstructBLIP and ties with Nullu on mPLUG-Owl2.

\begin{table}[t]
\centering
\small
\setlength{\tabcolsep}{5.2pt}
\renewcommand{\arraystretch}{1.08}
\caption{MM-Vet results across three LVLM backbones. We report scores on recognition (Rec.), OCR, knowledge (Know.), generation (Gen.), spatial reasoning (Spat.), math, and the overall total score. Higher scores indicate better performance.}
\begin{tabular}{llccccccc}
\toprule
\textbf{Backbone} & \textbf{Method}
& \textbf{Rec.} & \textbf{OCR} & \textbf{Know.}
& \textbf{Gen.} & \textbf{Spat.} & \textbf{Math} & \textbf{Total} \\
\midrule
\multirow{2}{*}{LLaVA-1.5}
& Vanilla & 34.0 & 22.6 & 16.1 & \textbf{20.1} & 24.4 & \textbf{11.5} & 29.5 \\
& ResOT     & \textbf{34.6} & \textbf{23.2} & \textbf{16.3} & 19.4 & \textbf{27.1} & \textbf{11.5} & \textbf{31.0} \\
\midrule
\multirow{2}{*}{mPLUG-Owl2}
& Vanilla & \textbf{38.7} & 27.0 & \textbf{23.8} & \textbf{25.4} & 29.5 & \textbf{3.8} & 34.3 \\
& ResOT     & 38.2 & \textbf{28.2} & 23.6 & 24.9 & \textbf{30.9} & \textbf{3.8} & \textbf{34.5} \\
\midrule
\multirow{2}{*}{InstructBLIP}
& Vanilla & 30.9 & 15.9 & 15.4 & 16.0 & 17.3 & \textbf{8.1} & 25.8 \\
& ResOT     & \textbf{33.5} & \textbf{17.1} & \textbf{19.6} & \textbf{18.0} & \textbf{18.1} & 7.3 & \textbf{27.7} \\
\bottomrule
\end{tabular}
\label{tab:mmvet_results}
\end{table}
\paragraph{Results on MM-Vet.}
ResOT improves the overall MM-Vet score across all three backbones (Table~\ref{tab:mmvet_results}). Performance varies across individual skills, but the overall score consistently increases for every model. Together with the hallucination results, this suggests that ResOT reduces hallucination without broadly degrading multimodal capability.

\begin{table*}[t]
\centering
\caption{Results on MMMU, TextVQA, ScienceQA, and MMBench across three LVLM backbones. Higher values are better.}
\label{tab:multimodal_benchmarks}
\small
\setlength{\tabcolsep}{6pt}
\renewcommand{\arraystretch}{0.92}
\begin{tabular}{@{}llcccc@{}}
\toprule
Model & Method & MMMU & TextVQA & ScienceQA & MMBench \\
\midrule

\multirow{3}{*}{LLaVA-1.5}
& Vanilla & 35.22 & 45.87 & 65.25 & \textbf{61.76} \\
& Nullu   & 34.89 & 43.04 & 64.95 & 59.29 \\
& ResOT   & \textbf{35.89} & \textbf{46.15}
          & \textbf{65.49} & \textbf{61.76} \\
\midrule

\multirow{3}{*}{mPLUG-Owl2}
& Vanilla & 36.56 & \textbf{55.36} & 68.07 & 63.62 \\
& Nullu   & 36.11 & 55.04 & 68.02 & 63.24 \\
& ResOT   & \textbf{36.67} & 55.17
          & \textbf{68.22} & \textbf{63.85} \\
\midrule

\multirow{3}{*}{InstructBLIP}
& Vanilla & 31.67 & \textbf{33.25} & 54.19 & 26.16 \\
& Nullu   & 31.67 & 31.11 & 54.27 & 24.73 \\
& ResOT   & \textbf{33.11} & 33.01
          & \textbf{54.83} & \textbf{27.32} \\
\bottomrule
\end{tabular}
\end{table*}
\paragraph{Evaluation on Additional Multimodal Benchmarks.}
Across MMMU, TextVQA, ScienceQA, and MMBench, ResOT consistently outperforms Nullu across all three backbones (Table~\ref{tab:multimodal_benchmarks}). Compared with the vanilla models, ResOT improves performance in 9 of the 12 settings and matches the baseline in one, with only modest decreases on TextVQA for mPLUG-Owl2 and InstructBLIP. Overall, the results show that ResOT preserves or improves general multimodal performance while reducing hallucination.

\subsection{Complexity Analysis}
\begin{table}[t]
\caption{Inference efficiency on LLaVA-1.5 with 128 generated tokens. Changes are computed against each method's paired Vanilla run.}
\label{tab:inference_efficiency}
\centering
\small
\setlength{\tabcolsep}{6pt}
\renewcommand{\arraystretch}{1.10}
\begin{tabular}{lccc}
\toprule
\textbf{Method}
& \textbf{Tokens/s (vs. Vanilla)} $\uparrow$
& \textbf{Peak Memory Increase (MiB)} $\downarrow$
& \textbf{Preserves LVLM Weights} \\
\midrule
OPERA     & 0.752 ($-97.87$\%) & 25219.631 & \cmark \\
VCD       & 17.462 ($-52.64$\%)& 724.160   & \cmark \\
HALC      & 1.858 ($-95.21$\%) & 4728.521  & \cmark \\
ICT       & 24.285 ($-39.46$\%)& 0.250     & \cmark \\
VTI       & 29.940 ($-24.77$\%)& 56.504    & \cmark \\
VISTA     & 22.200 ($-51.43$\%)& 3761.718  & \cmark \\
Nullu     & 45.850 ($-1.64$\%) & 0.000     & \xmark \\
HulluEdit & 17.843 ($-56.16$\%)& 112.595   & \cmark \\
\midrule
\textbf{ResOT}
          & \textbf{40.071 ($-18.76$\%)}
          & \textbf{8.147}
          & \textbf{\cmark} \\
\bottomrule
\end{tabular}
\end{table}
ResOT estimates the residual subspace and Gaussian OT map offline, while online inference only applies the transport maps to the current token hidden representations. Let $L$ denote the number of intervened layers. At each layer, projection into the residual subspace and reconstruction in the original representation space cost $O(dk)$. Evaluating the affine transport map and the Gaussian likelihood ratio costs $O(k^2)$, while correction clipping costs $O(k)$. The total additional computation per generated token is therefore $O(L(dk+k^2))$. Since $k \ll d$, the dominant term is $O(Ldk)$. Storing the residual bases, transport maps, and Gaussian statistics requires $O(L(dk+k^2))$ memory.

On LLaVA-1.5, ResOT retains 81.24\% of the throughput of its paired vanilla run and increases peak GPU memory by 8.15 MiB (Table~\ref{tab:inference_efficiency}). All efficiency measurements in Table~\ref{tab:inference_efficiency} were conducted under the same hardware configuration. The additional state is stored separately from the model weights, so the intervention can be enabled or disabled without modifying the backbone checkpoint.

\section{Sensitivity and Ablation Analysis}
\label{app:sens_and_abla_analysis}
\subsection{Quantitative Analysis of Representations}
\label{app:quantitative_analysis_of_representations}

For the analyses in this subsection, we fit the subspaces and transport maps on 4,000 paired descriptions and evaluate them on the remaining 1,000 pairs. All visualizations and quantitative results are computed on the held-out split.

\paragraph{Visualization across Layers.}
\begin{figure}[t]
\centering
\includegraphics[width=0.9\linewidth]{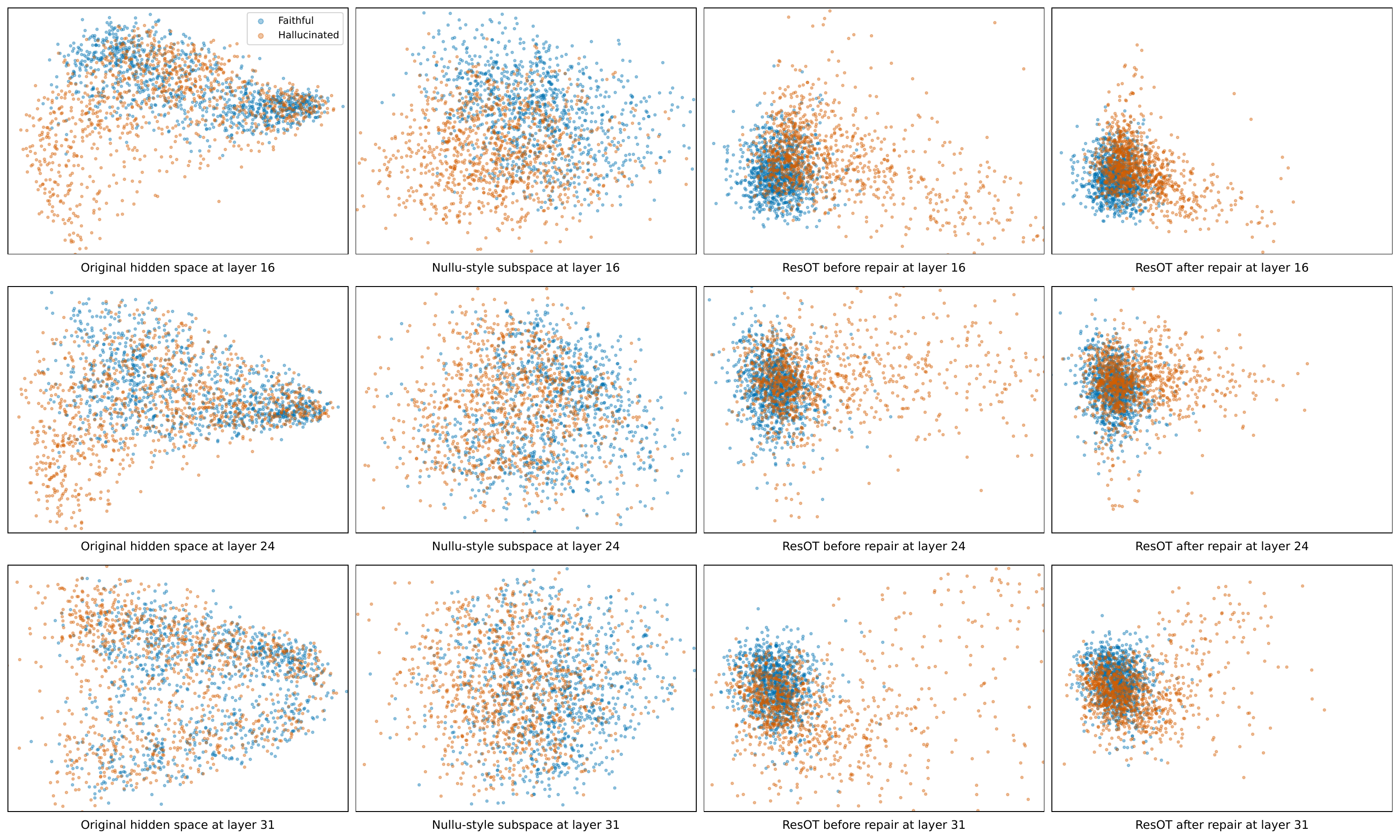}
\caption{
\textbf{Representation visualizations across layers.} Rows correspond to layers 16, 24, and 31. Columns show the original representation space, the Nullu-style subspace, and the ResOT residual subspace before and after repair. We use PCA for visualization.
}
\label{fig:projection_scatter_layers}
\end{figure}
Figure~\ref{fig:projection_scatter_layers} extends the representation visualization to layers 16, 24 and 31. Across all three layers, we observe a consistent pattern: the residual subspace makes the difference between faithful and hallucinated representations more apparent, while Gaussian OT moves the hallucinated distribution toward the faithful one. This shows that the representation pattern observed at layer 31 is not specific to a single layer.

\begin{figure}[t]
\centering
\begin{subfigure}[t]{0.24\linewidth}
    \centering
    \includegraphics[width=\linewidth]{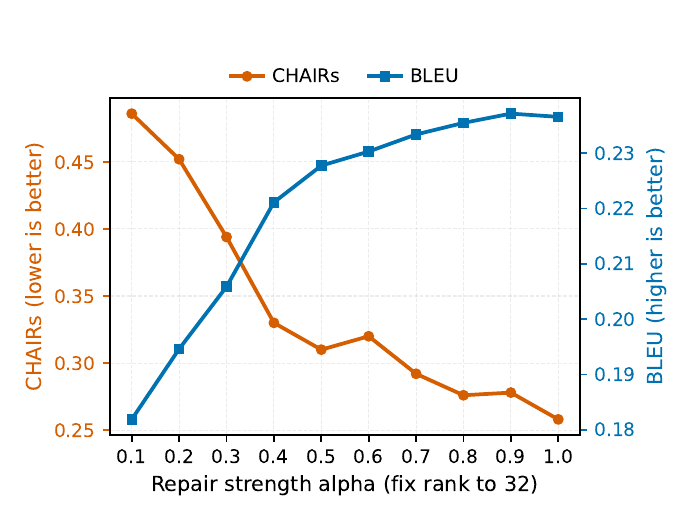}
    \caption{CHAIR vs. $\alpha$}
    \label{fig:chair_alpha}
\end{subfigure}\hfill
\begin{subfigure}[t]{0.24\linewidth}
    \centering
    \includegraphics[width=\linewidth]{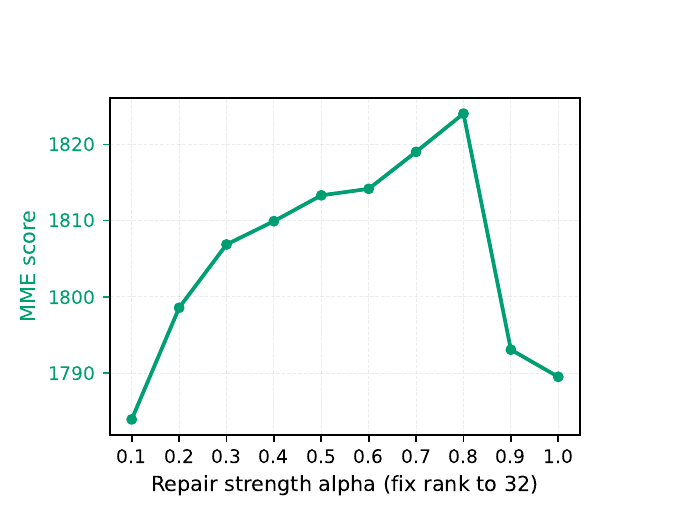}
    \caption{MME vs. $\alpha$}
    \label{fig:mme_alpha}
\end{subfigure}\hfill
\begin{subfigure}[t]{0.24\linewidth}
    \centering
    \includegraphics[width=\linewidth]{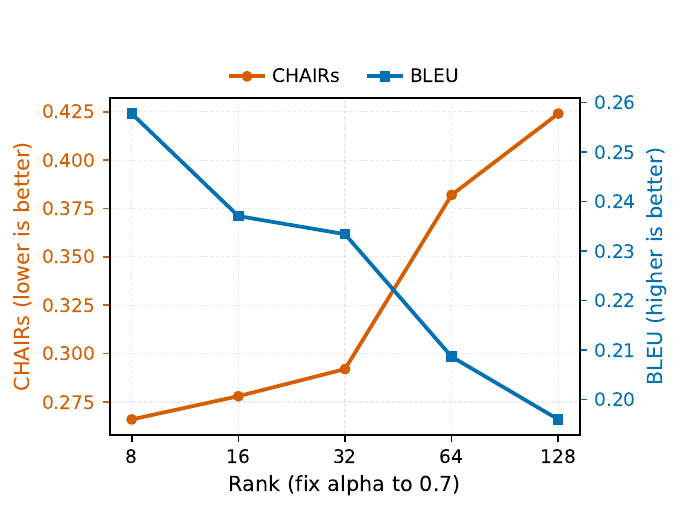}
    \caption{CHAIR vs. rank}
    \label{fig:chair_rank}
\end{subfigure}\hfill
\begin{subfigure}[t]{0.24\linewidth}
    \centering
    \includegraphics[width=\linewidth]{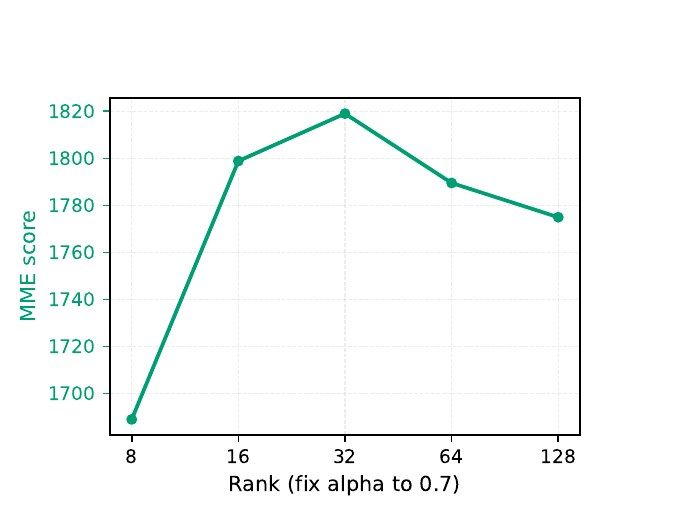}
    \caption{MME vs. rank}
    \label{fig:mme_rank}
\end{subfigure}
\vspace{-4pt}
\caption{
Sensitivity analysis of the repair strength $\alpha$ and residual subspace rank on LLaVA-1.5. For $\alpha$ sensitivity, we fix the rank to 32. For rank sensitivity, we fix $\alpha$ to 0.7.
}
\label{fig:sensitivity}
\vspace{-8pt}
\end{figure}

\paragraph{Subspace Separability.}
\begin{table}[t]
\centering
\caption{
Separability between faithful and hallucinated LLaVA-1.5 representations in the subspace on the evaluation split. Higher values indicate stronger separability.
}
\label{tab:separability}
\begin{tabular}{clccc}
\toprule
Layer & Setting & Mahalanobis Dist. $\uparrow$
& LDA Acc. $\uparrow$ & LDA AUC $\uparrow$ \\
\midrule
\multirow{2}{*}{16}
& Nullu-style
& 1.5033 & 0.7795 & 0.8674 \\
& ResOT before repair
& \textbf{2.4725} & \textbf{0.8865} & \textbf{0.9544} \\
\midrule
\multirow{2}{*}{24}
& Nullu-style
& 1.3405 & 0.7590 & 0.8366 \\
& ResOT before repair
& \textbf{2.1708} & \textbf{0.8565} & \textbf{0.9351} \\
\midrule
\multirow{2}{*}{31}
& Nullu-style
& 1.2530 & 0.7325 & 0.8161 \\
& ResOT before repair
& \textbf{1.9297} & \textbf{0.8460} & \textbf{0.9149} \\
\bottomrule
\end{tabular}
\vspace{-8pt}
\end{table}
To quantify the localization achieved by subspace construction, we compare the separability of faithful and hallucinated representations in the Nullu-style subspace and the ResOT residual space before repair. As shown in Table~\ref{tab:separability}, the residual space consistently yields a higher Mahalanobis distance, LDA accuracy, and LDA AUC across all evaluated layers, demonstrating that it better isolates hallucination-related deviations.

\paragraph{Center Alignment by Gaussian OT.}
\begin{table}[t]
\centering
\caption{
Center-distance analysis across layers. $F$, $H$, and $R$ denote faithful, hallucinated, and repaired representations, respectively.
}
\label{tab:center_distance}
\begin{tabular}{lccc}
\toprule
Layer & Nullu ($F \rightarrow H$) & Before repair ($F \rightarrow H$) & After repair ($F \rightarrow R$) \\
\midrule
16 & 1.9758 & 1.2845 & \textbf{0.0537} \\
24 & 4.0293 & 2.9075 & \textbf{0.1265} \\
31 & 4.4187 & 2.5757 & \textbf{0.0994} \\
\bottomrule
\end{tabular}
\end{table}
Table~\ref{tab:center_distance} further quantifies the repair effect. Faithful and hallucinated centers are clearly separated in the residual subspace before repair. After applying Gaussian OT, the distance between the faithful and repaired centers is significantly reduced.

\paragraph{Selectivity of Representation Repair.}
\begin{table}[!htbp]
\centering
\small
\setlength{\tabcolsep}{8pt}
\renewcommand{\arraystretch}{1.08}
\caption{
Average relative displacement of faithful and hallucinated samples in the representation space. H/F denotes the ratio of hallucinated to faithful displacement. The 16--31 row reports the average across all selected layers.
}
\label{tab:relative_displacement}
\begin{tabular}{lccc}
\toprule
\textbf{Layer}
& \textbf{Faithful (\%)}
& \textbf{Hallucinated (\%)}
& \textbf{H/F} \\
\midrule
16 & 1.5903 & 7.1353 & 4.49$\times$ \\
24 & 1.1953 & 5.9745 & 5.00$\times$ \\
31 & 1.3840 & 4.2196 & 3.05$\times$ \\
\midrule
16--31 & 1.3425 & 6.0281 & 4.49$\times$ \\
\bottomrule
\end{tabular}
\end{table}
For each class $C \in \{F, H\}$, we measure the average relative displacement of each sample in the full latent space:
\begin{equation}
    D_C^{\ell}=\frac{1}{N_C}\sum_{i=1}^{N_C}\frac{\left\| h_{C,i}^{\ell,\mathrm{after}} - h_{C,i}^{\ell,\mathrm{before}}\right\|_2}{\left\|h_{C,i}^{\ell,\mathrm{before}}\right\|_2}.
\end{equation}
Here, $N_C$ is the number of samples in class $C$, and the hidden states are measured before and after repair. For this analysis, we set $\alpha=1$. Table~\ref{tab:relative_displacement} shows that hallucinated representations exhibit larger relative displacements than faithful representations across layers. This difference shows that ResOT applies the repair selectively rather than shifting all representations uniformly.

\subsection{Hyperparameter Sensitivity}
Figure~\ref{fig:sensitivity} analyzes the sensitivity of ResOT to the repair strength $\alpha$ and residual subspace rank on LLaVA-1.5. With the rank fixed at 32, increasing $\alpha$ generally lowers $\text{C}_s$ and improves BLEU-1, while MME peaks at $\alpha=0.8$ and drops beyond this point. Smaller ranks favor $\text{C}_s$ and BLEU-1, whereas ranks above 32 weaken all three metrics. The sensitivity analysis shows that larger repair strength is not uniformly beneficial and that overly large residual ranks can degrade performance.

\subsection{Gaussian Diagnostics in the Residual Subspace}
ResOT models faithful and hallucinated residual representations with Gaussian distributions to obtain a closed-form transport map. We therefore examine how well this approximation describes the residual representations used to estimate the OT maps. We use the same 5,000 calibration pairs after hyperparameter selection. At each layer, faithful and hallucinated residual vectors are whitened separately using their estimated means and covariances. For the marginal analysis, we aggregate the whitened coordinate values across samples and dimensions and compare them with $\mathcal{N}(0,1)$. To examine the joint distribution, we compute the squared Mahalanobis distance of each residual vector and compare its distribution with $\chi_k^2$, where $k$ is the residual dimension.

\begin{figure*}[t]
    \centering
    \includegraphics[width=0.9\textwidth]{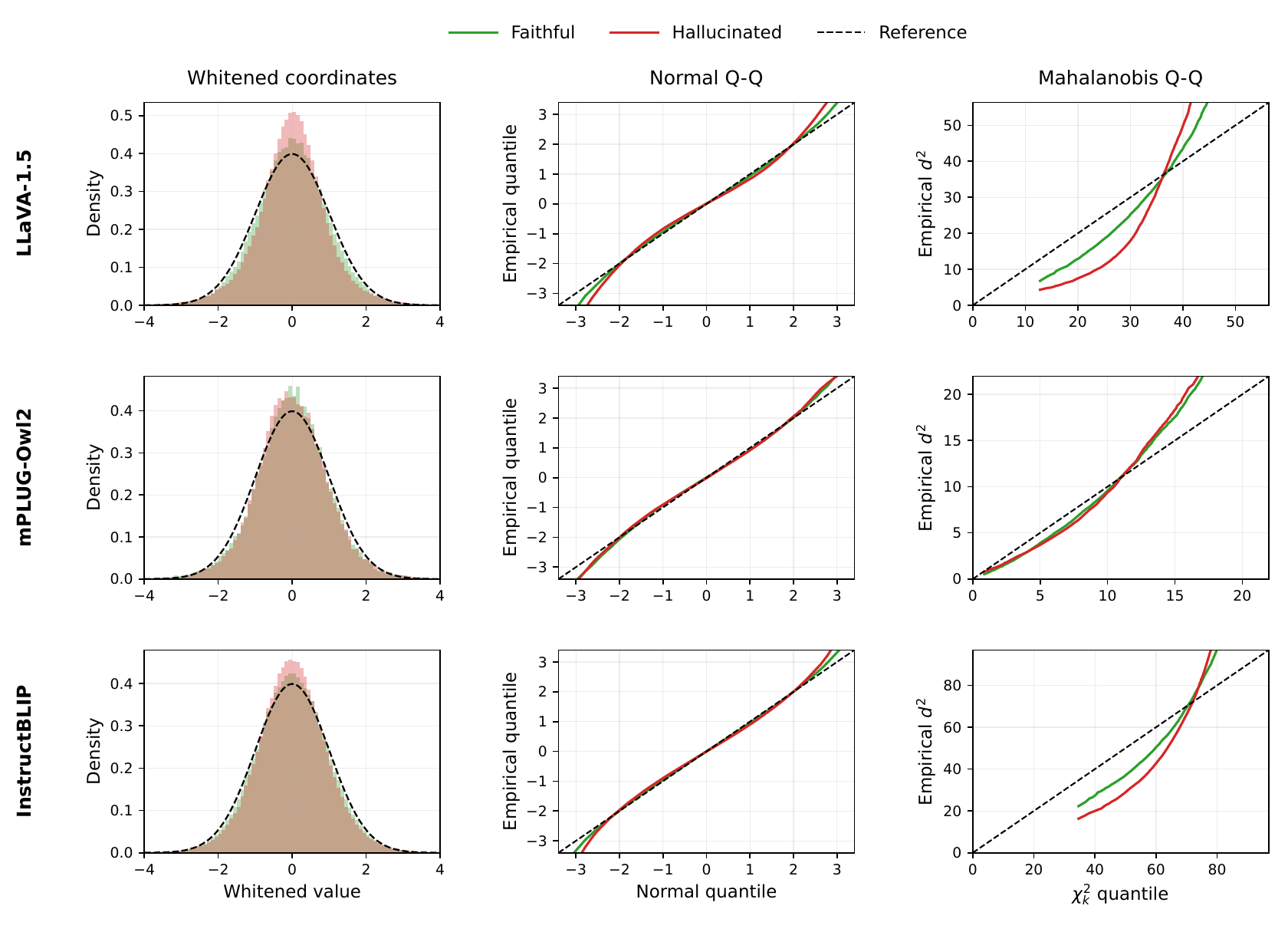}
    \caption{
    \textbf{Gaussian diagnostics at layer 31.} Columns show histograms of whitened coordinates (left), normal Q--Q plots (middle), and Mahalanobis Q--Q plots (right). Green and red denote faithful and hallucinated representations, respectively. Dashed black lines show the standard normal reference in the histograms and $y=x$ in the Q--Q plots.
    }
    \label{fig:layer31_gaussian_diagnostics}
    \vspace{-8pt}
\end{figure*}

Figure~\ref{fig:layer31_gaussian_diagnostics} provides a detailed view at layer 31. The whitened marginal distributions broadly follow the standard normal reference, and the normal Q--Q plots remain close to the diagonal over much of the central range. Larger deviations appear toward the tails. The Mahalanobis Q--Q plots show more noticeable departures from the $\chi_k^2$ reference, indicating that the joint residual distributions are not exactly multivariate Gaussian. Thus, the marginal coordinates are reasonably approximated by Gaussian distributions in their central regions, while the full joint distributions retain structure beyond a Gaussian model.

\begin{figure*}[t]
    \centering
    \includegraphics[width=0.8\textwidth]{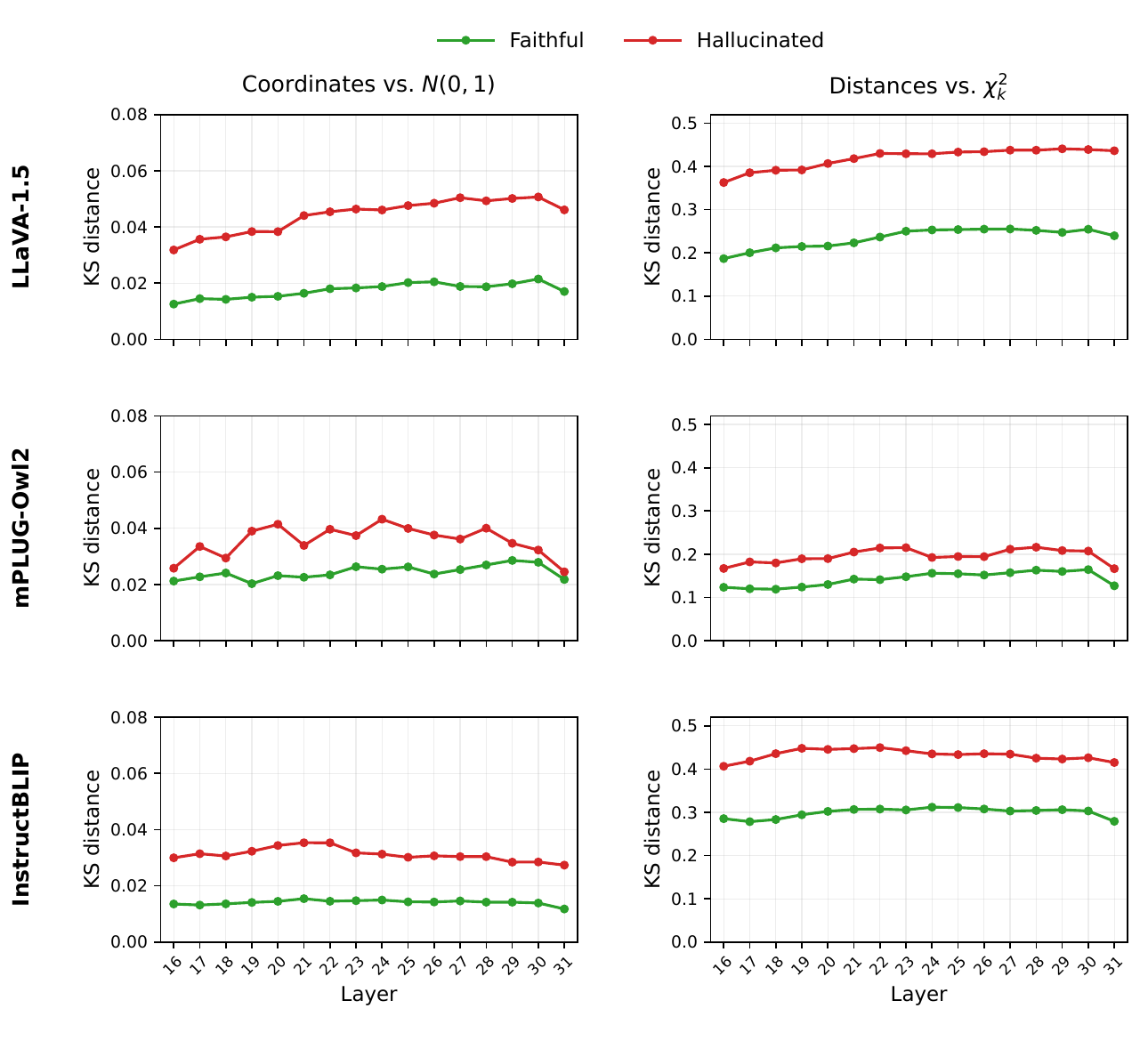}
    \caption{
    \textbf{Gaussian diagnostics across layers 16--31.}
    Columns report the KS distance of whitened coordinate values to $\mathcal{N}(0,1)$ (left) and the KS distance of squared Mahalanobis distances to $\chi_k^2$ (right).
    Green and red curves denote faithful and hallucinated representations, respectively.
    }
    \label{fig:gaussian_summary_all_backbones}
    \vspace{-8pt}
\end{figure*}

Figure~\ref{fig:gaussian_summary_all_backbones} extends the diagnostic across all repaired layers. The marginal KS distances remain relatively small and stable across layers, whereas the Mahalanobis KS distances are generally larger. The same pattern appears across all three backbones, suggesting that the behavior observed at layer 24 is not specific to a single layer or model.

Overall, these diagnostics support using a Gaussian model as a compact approximation of the residual distributions rather than as an exact description of their full shape. ResOT uses the estimated means and covariances to construct the transport map, while the observed departures from Gaussianity are concentrated more strongly in the joint distribution and its tails.

\section{Limitation}
ResOT uses a low-dimensional Gaussian model to define the transport target. This model can capture differences in mean and covariance but does not describe all higher-order structure in the residual distributions. In addition, calibration uses representations from complete descriptions, whereas repair is applied to a partially generated context. This potential distributional shift remains insufficiently explored. Our evaluation focuses primarily on object hallucination. The applicability of ResOT to other forms of multimodal error remains to be verified.

\end{document}